\documentclass{article}

\usepackage{arxiv}

\usepackage[utf8]{inputenc} 
\usepackage[T1]{fontenc}    
\usepackage{hyperref}       
\usepackage{url}            
\usepackage{booktabs}       
\usepackage{amsfonts}       
\usepackage{nicefrac}       
\usepackage{microtype}      
\usepackage{lipsum}
{}
{}
{}

\usepackage{booktabs} 
\usepackage{pifont}
\newcommand{\xmark}{\ding{55}}%
\usepackage[T1]{fontenc}

\usepackage{lmodern}

\usepackage{xcolor}
\usepackage{multirow, array} 
\usepackage{tabularx}
\usepackage{booktabs}
\usepackage{colortbl}
\usepackage{graphicx}
\usepackage{amsthm}
\usepackage{soul}
\usepackage{url}
\definecolor{Gray}{gray}{0.85}
\newcommand{\virgolette}[1]{``#1''}
\usepackage{float}
\floatstyle{ruled}
\newfloat{program}{thp}{lop}
\floatname{program}{Listing}
\usepackage{amssymb}

\usepackage[ruled]{algorithm2e} 

\title{Survey on Visual Sentiment Analysis}

\author{\small
  Alessandro Ortis \\
  University of Catania\\
  Viale A. Doria, 6 Catania - 95125 \\
  \texttt{ortis@dmi.unict.it} \\
   \And
   Giovanni Maria Farinella\\
University of Catania\\
Viale A. Doria, 6 Catania - 95125 \\
\texttt{gfarinella@dmi.unict.it} \\
\And
  Sebastiano Battiato \\
University of Catania\\
Viale A. Doria, 6 Catania - 95125 \\
\texttt{battiato@dmi.unict.it} \\
}

\begin{document}
\maketitle

{\small \emph{This paper is a postprint of a paper accepted by IET Image Processing and is subject to Institution of Engineering and Technology Copyright. When the final version is published, the copy of record will be available at the IET Digital Library. DOI: 10.1049/iet-ipr.2019.1270\\}}

\begin{abstract}
	Visual Sentiment Analysis aims to understand how images affect people, in terms of evoked emotions. 
Although this field is rather new, over the last years, a broad range of techniques have been developed for various data sources and problems, resulting in a large body of research. 
This paper reviews pertinent publications and tries to present an  exhaustive e view of the field. After a description of the task and the related applications, the subject is tackled under different main headings. 
The paper also describes principles of design of general Visual Sentiment Analysis systems from three main points of view: emotional models, dataset definition, feature design.
A formalization of the problem is discussed, considering different levels of granularity, as well as the components that can affect the sentiment toward an image in different ways. To this aim, this paper considers a structured formalization of the problem which is usually used for the analysis of text, and discusses it's suitability in the context of Visual Sentiment Analysis. 
The paper also includes a description of new challenges, the evaluation from the viewpoint of progress toward more sophisticated systems and related practical applications, as well as a summary of the insights resulting from this study.
\end{abstract}


\section{Introduction and Motivations}
Nowadays, the amount of public available information encourages the study and development of Sentiment Analysis algorithms that analyse huge amount of users' data with the aim to infer reactions about topics, opinions, trends to understand the mood of the users who produce and share information through the web. The aim of Sentiment Analysis is to extract the attitude of people toward a topic or the intended emotional affect the author wishes to have on the readers. The tasks of this research field are challenging as well as very useful in practice. Sentiment analysis finds several practical applications, since opinions influence many human decisions either in business and social activities.  
As instance, companies are interested in monitoring people opinions toward their products or services, as well as customers rely on feedbacks of other users to evaluate a product before they purchase it.
With the growth of social media (i.e., reviews, forums, blogs and social networks), individuals and organizations are increasingly using public opinions  for their decision making~\cite{liu2012survey}.

The basic task in Sentiment Analysis is the polarity classification of an input text (e.g., taken from a review, a comment or a social post) in terms of positive, negative or neutral polarity. This analysis can be performed at document, sentence or feature level. The methods of this area are useful to capture public opinion about products, services, marketing, political preferences and social events. For example the analysis of the activity of Twitter's users can help to predict the popularity of parties or coalitions. The achieved results in Sentiment Analysis within micro-blogging have shown that Twitter posts reasonably reflect the political landscape~\cite{tumasjan2010predicting}.
Historically, Sentiment Analysis techniques have been developed for the analysis of text~\cite{pang2008opinion}, whereas limited efforts have been employed to extract (i.e., infer) sentiments from visual contents~(e.g., images and videos). 
Even though the scientific research has already achieved notable results in the field of textual Sentiment Analysis in different contexts (e.g., social network posts analysis, product reviews, political preferences, etc.)~\cite{zhang2018deep,dohaiha2018deep}, the task to understand the mood from a text has several difficulties given by the inherent ambiguity of the various languages (e.g., ironic sentences), cultural factors, linguistic nuances and the difficulty of generalize any text analysis solution to different language vocabularies. The different solutions in the field of text Sentiment Analysis have not yet achieved a level of reliability good enough to be implemented without enclosing the related context. For example, despite the existence of natural language processing tools for the English language, the same tools cannot be used directly to analyse text written in other languages.
Although NLP (Natural Language Processing) and information retrieval researchers proposed several approaches to address the problem of Sentiment Analysis~\cite{pang2008opinion}, the social media context offers some additional challenges. Beside the huge amounts of available data, typically the textual communications on social networks consist of short and colloquial messages. Moreover, people tend to use also images and videos, in addition to the textual messages, to express their experiences through the most common social platforms.
The information contained in such visual contents are not only related to semantic contents such as objects or actions about the acquired picture, but also cues about affect and sentiment conveyed by the depicted scene. Such information is hence useful to understand the emotional impact (i.e., the evoked sentiment) beyond the semantic.
For these reasons images and videos have become one of the most popular media by which people express their emotions and share their experiences in the social networks, which have assumed a crucial role in collecting data about people's opinions and feelings. 
The images shared in social media platforms reflect visual aspects of users' daily activities and interests. Such growing user generated images represent a recent and powerful source of information useful to analyse users' interests. 
Extracting the emotions inherently underlying images perceived by the viewers could promote new approaches to several application fields such as brand perception, assessment of customer satisfaction, advertising, media analytic, etc.
The diffusion of personal mobile devices constantly connected to Internet services (e.g., smartphones) and the growth of social media platforms introduced a new communication paradigm by which people that share multimedia data. 
In this context, uploading images to a social media platform is the new way by which people share their opinions and experiences.  
This provides a strong motivation for research on this field, and offers many challenging research problems.
The proposed paper aims to introduce this emerging research field. It analyses the related problems, provides a brief overview of current research progress, as well as discusses the major issues and outline the new opportunities and challenges in this area.	
This paper extends our preliminary work~\cite{sigmap19ortis}, by significantly augmenting the number and type of reviewed papers, as well as the level of details, including publications that address specific aspects of the issues mentioned in~\cite{sigmap19ortis} related to features, models, datasets and challenges. This paper further adds a detailed study of the works that tackle specific tasks (e.g., popularity and popularity dynamics, relative attributes, virality, etc.).
\\ 
Besides the overview of the state of the art, the paper proposes a critical viewpoint on the previous works considering the employed features, datasets and methods. Indeed, Section~\ref{stateoftheart} presents a review of the state of the art works that mostly influenced the research field. In particular, an overview of the most significant works in the field of Visual Sentiment Analysis published between 2010 and 2019 is presented. The literature is presented in a chronological order, highlighting similarities and differences between the different works, with the aim to drive the reader along the evolution of the developed methods. 
Then, a structured and critical dissertation of the topics and issues that a researcher interested in Visual Sentiment Analysis needs to address is presented in Section~\ref{systems}. These include the choice of the emotional categories, the dataset definition~(e.g., existing datasets, sources of data, labelling, size of a good dataset, etc.), as well as considerations about the features that may influence the sentiment toward an image. It provides a complete overview of the system design choices, with the aim to provide a deep debate about each specific issue related to the design of a Visual Sentiment Analysis systems: emotions representation schemes, existing datasets and features. Each aspect is discussed, and the different possibilities are compared one each other, with proper references to the state of the art.  
Section~\ref{problem} provides a thorough formalization of the problem, by abstracting all the components that could affect the sentiment associated to an image, including the sentiment holder and the time factor, often ignored by the existing methods. References to the state of the art addressing each component are reported, as well as the different alternative solutions are proposed. 
The paper also addresses the future perspectives of the field by describing five emerging tasks related to Visual Sentiment Analysis, with proper references, in Section~\ref{challenges}. The paper introduces some additional challenges and techniques that could be investigated, proposing suggestions for new methods, features and datasets.
In this sense, the proposed paper aims to serve as a guide for research scientist that are interested in Visual Sentiment Analysis. Indeed, this paper is intended to be a reference source of contents in the field, by providing a structured description and analysis of the past works, a detailed presentation of the different existing datasets with pros and cons, as well as a perspective on current challenges and for future directions. 
Section~\ref{discussion} reports the insights as outcome of the proposed study. Then, Section~\ref{conclusions} concludes the paper by providing a summary of the previous sections and by suggesting directions for future works. 

\section{State of the Art}\label{stateoftheart}
Visual Sentiment Analysis is a recent research area.
Most of the works in this new research field rely on previous studies on emotional semantic image retrieval~\cite{colombo1999semantics, schmidt2009collective, wei2006image, zhao2014affective}, which make connections between low-level image features and emotions with the aim to perform automatic image retrieval and categorization. These works have been also influenced by empirical studies from psychology and art theory~\cite{bradley1994emotional,itten1973art, lang1993network, osgood1952nature, valdez1994effects, russell1977evidence}. 
Other research fields close to Visual Sentiment Analysis are those considering the analysis of the image aesthetic~\cite{datta2006studying, joshi2011aesthetics, joshi2014aesthetics, marchesotti2011assessing,Ravi2012}, interestingness~\cite{isola2011makes}, affect~\cite{jia2012can} and popularity~\cite{gelli2015image,khosla2014makes,mcparlane2014nobody,totti2014impact}.


The first paper on Visual Sentiment Analysis aims to classify images as \virgolette{positive} or \virgolette{negative} and dates back on 2010~\cite{siersdorfer2010analyzing}.
In this work the authors studied the correlations between the sentiment of images and their visual content. 
They assigned numerical sentiment scores to each picture based on their accompanying text (i.e., meta-data). To this aim, the authors used the  SentiWordNet~\cite{esuli2006sentiwordnet} lexicon to extract sentiment score values from the text associated to images.
This work revealed that there are strong correlations between sentiment scores extracted from Flickr meta-data (e.g., image title, description and tags provided by the user) and visual features (i.e., SIFT based bag-of-visual words, and local/global RGB histograms). 

In~\cite{machajdik2010affective} a study on the features useful to the task of affective classification of images is presented. The insights from the experimental observation of emotional responses with respect to colors and art have been exploited to empirically select the image features. 
To perform the emotional image classification, the authors considered the 8 emotional output categories as defined in~\cite{mikels2005emotional} (i.e., Awe,  Anger, Amusement, Contentment, Excitement, Disgust, Sad, and Fear). 


In~\cite{borth2013large} the authors built a large scale Visual Sentiment Ontology (VSO) of semantic concepts based on psychological theories and web mining (SentiBank). A concept is expressed as an adjective-noun combination called Adjective Noun Pair (ANP) such as \virgolette{beautiful flowers} or \virgolette{sad eyes}. After building the ontology consisting of 1.200 ANP, they trained a set of 1.200 visual concept detectors which responses can be exploited as a sentiment representation for a given image. Indeed, the 1.200 dimension ANP outputs (i.e., the outputs of the ANP detectors) can be exploited as features to train a sentiment classifier. To perform this work the authors extracted adjectives and nouns from videos and images tags retrieved from YouTube and Flickr respectively. These images and videos have been searched using the words corresponding to the 24 emotions defined in the Plutchik Wheel of Emotion~\cite{plutchik1980general}, a well known psychological model of human emotions. The authors released a large labelled image dataset composed by half million Flickr images regarding to 1.200 ANPs.  
Results show that the approach based on SentiBank concepts outperforms text based method in tweet sentiment prediction experiments. 
Furthermore, the authors compared the SentiBank representation with shallow features (colour histogram, GIST, LBP, BoW) to predict the sentiment reflected in images. To this end, they used two different classification models (LinearSVM and Logistic Regression) achieving significant performance improvements when using the SentiBank representation.
The proposed mid-level representation has been further evaluated in the emotion classification task considered in~\cite{machajdik2010affective}, obtaining better results. 

In 2013, Yuan et al.~\cite{yuan2013sentribute} employed scene-based attributes to define mid-level features, and built a binary sentiment classifier on top of them. 
Furthermore, their experiments demonstrated that adding a facial expression recognition step helps the sentiment prediction task when applied to images with faces.

In 2014, Yang et al.~\cite{yang2014your} proposed a Sentiment Analysis approach based on a graphical model which is used to represent the connections between visual features and friends interactions (i.e., comments) related to the shared images.
The exploited visual features include saturation, saturation contrast, bright contrast, cool color ratio, figure-ground color difference, figure-ground area difference, background texture complexity, and foreground texture complexity. In this work the authors considered the Ekman's emotion model~\cite{ekman1987universals}.

Chen et al.~\cite{chen2014deepsentibank} introduced a CNN (Convolutional Neural Network) based approach, also known as \virgolette{SentiBank 2.0} or \virgolette{DeepSentiBank}. They performed a fine-tuning training on a CNN model previously trained for the task of object classification to classify images in one of a 2.096 ANP category (obtained by extending the previous SentiBank ontology~\cite{borth2013large}). This approach significantly improved the ANP detection with respect to~\cite{borth2013large}. Similarly to~\cite{borth2013large}, this approach provides a sentiment feature (i.e., a representation) of an image that can be exploited by further systems.

In contrast to the common task of infer the affective concepts intended by the media content publisher (i.e., by analysing the text associated to the image by the publisher), the method proposed in~\cite{chen2014predicting} tries to predict what concepts will be evoked to the image viewers.

In~\cite{xu2014visual} a pre-trained CNN is used as a provider of high-level attribute descriptors in order to train two sentiment classifiers based on Logistic Regression. Two types of activations are used as visual features, namely the fc7 and fc8 features (i.e., the activations of the seventh and eighth fully connected layers of the CNN respectively). 
The authors propose a fine-grained sentiment categorization, classifying the polarity of a given image through a 5-scale labelling scheme: \textit{\virgolette{strong negative}, \virgolette{weak negative}, \virgolette{neutral}, \virgolette{weak positive}, and \virgolette{strong positive}}.
The evaluation of this approach considers two baseline methods taken from the state of the art, namely low-level visual features and SentiBank, both introduced in~\cite{borth2013large}, in comparison with their approaches (the fc7 and fc8 based classifiers). The experimental setting evaluates all the considered methods on two real-world dataset related to Twitter and Tumblr, whose images have been manually labelled considering the above described 5-scale score scheme.
The results suggest that the methods proposed in~\cite{xu2014visual} outperform the baseline methods in visual sentiment prediction.

The authors of~\cite{you2015robust} proposed to use a progressive approach for training a CNN (called Progressive CNN or PCNN) in order to perform visual Sentiment Analysis in terms of \virgolette{positive} or \virgolette{negative} polarity. They first trained a CNN architecture with a dataset of half million Flickr images introduced in~\cite{borth2013large}. At training time, the method selects a subset of training images which achieve high prediction scores. Then, this subset is used to further fine-tune the obtained CNN. 
In the architecture design they considered a last fully connected layer with 24 neurons. This design decision has been taken with the aim to let the CNN learn the response of the 24 Plutchik's emotions~\cite{plutchik1980general}.
An implementation of the architecture proposed in~\cite{you2015robust} is publicly available. The results of experiments performed on a set of manually labelled Twitter images show that the progressive CNN approach obtain better results with respect to other previous algorithms, such as~\cite{borth2013large} and~\cite{yuan2013sentribute}. 

\begin{figure}[t]
	\centering
	\includegraphics[width=0.6\linewidth]{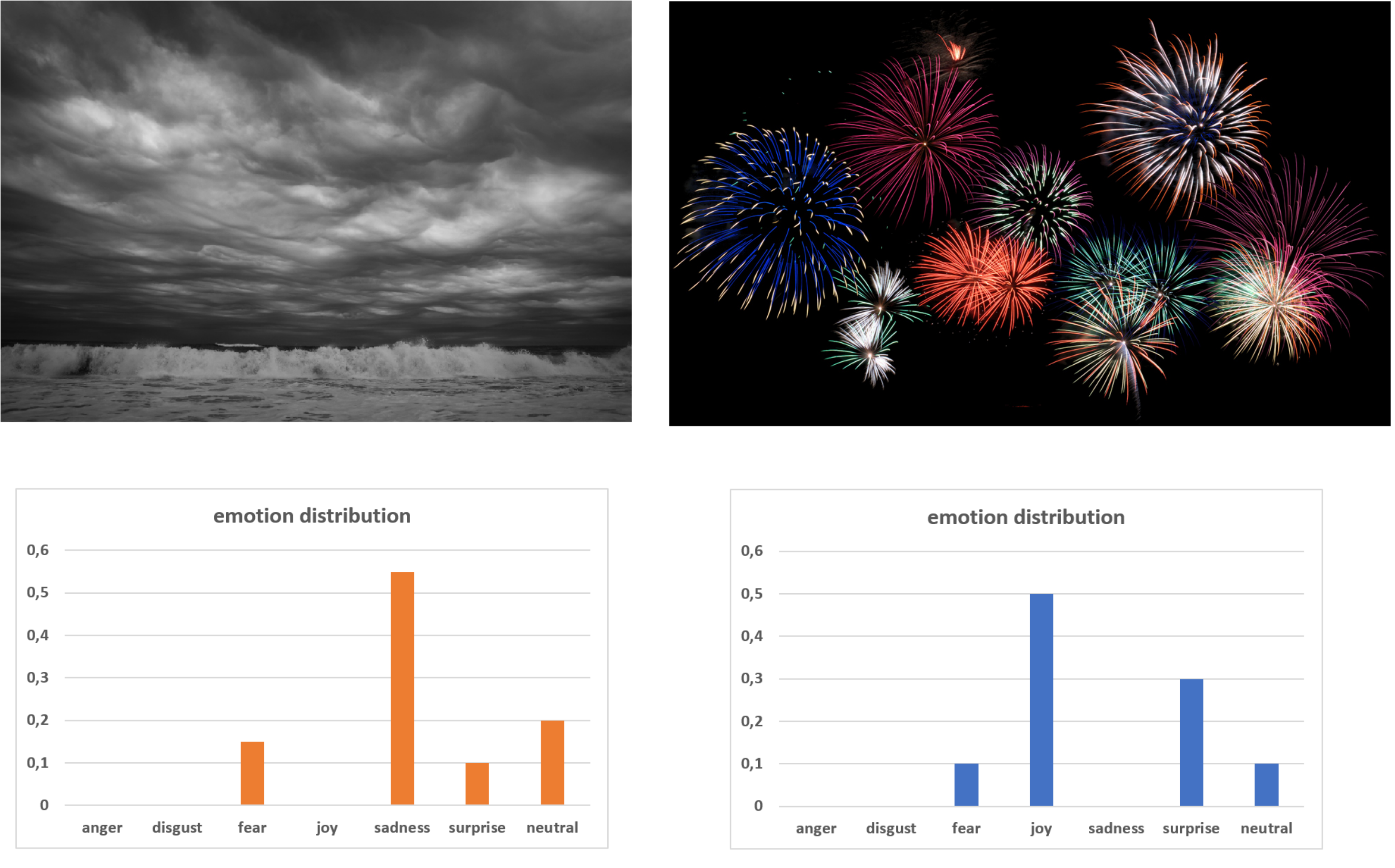}
	\caption{Examples of image emotion distributions.}
	\label{figDistribution}
\end{figure}

Considering that the emotional response of a person viewing an image may include multiple emotions, the authors of~\cite{peng2015mixed} aimed to predict a distribution representation of the emotions rather than a single dominant emotion from (see Figure~\ref{figDistribution}). The authors compared three methods to predict such emotion distributions: a Support Vector Regressor (based on hand crafted features related to edge, color, texture, shape and saliency), a CNN for both classification and regression. They also proposed a method to change the evoked emotion distribution of an image by editing its texture and colors. Given a source image and a target one, the proposed method transforms the color tone and textures of the source image to those of the target one. 
The result is that the edited image evokes emotions closer to the target image than the original one. This approach has been quantitatively evaluated by using four similarity measures between distributions.
For the experiments, the authors consider a set of 7 emotion categories, corresponding to the 6 basic emotions defined by Ekman in~\cite{ekman1987universals} and the neutral emotion. Furthermore, the authors proposed a sentiment database called~Emotion6. The experiments on evoked emotion transfer suggest that holistic features such as the color tone can influence the evoked emotion, albeit the emotion related to images with high level semantics are difficult to be shaped according to an arbitrary target image.

In~\cite{wang2015unsupervised} the textual data, such as comments and captions, related to the images are considered as contextual information. Differently from the previous approaches, which exploit low-level features~\cite{jia2012can}, mid-level features~\cite{borth2013large, yuan2013sentribute} and Deep Learning architectures~\cite{you2015robust, peng2015mixed}, the framework in~\cite{wang2015unsupervised} implements an unsupervised approach~(USEA - Unsupervised SEntiment Analysis).
In~\cite{campos2015diving} a CNN pre-trained for the task of Object Classification is fine-tuned to accomplish the task of visual sentiment prediction. Then, with the aim to understand the contribution of each CNN layer for the task, the authors performed an exhaustive layer-per-layer analysis of the fine-tuned model. Indeed, the traditional approach consists in initializing the weights obtained by training a CNN for a specific task, and replacing the last layer with a new one containing a number of units corresponding to the number of classes of the new target dataset.
The experiments performed in this paper explored the possibility to use each layer as a feature extractor and training individual classifiers. This layer by layer study allows measuring the performance of the different layers which is useful to understand how the layers affect the whole CNN performances. Based on the layer by layer analysis, the authors proposed several CNN architectures obtained by either removing or adding layers from the original CNN.

Even though the conceptual meaning of an image is the same for all cultures, each culture may have a different sentimental expression of a given concept. 
Motivated by this observation, Jou et al.~\cite{jou2015visual} extended the ANP ontology defined in~\cite{borth2013large} for a multi-lingual context. Specifically, the method provides a multilingual sentiment driven visual concept detector in 12 languages. The resulting Multilingual Visual Sentiment Ontology (MVSO) provides a rich information source for the analysis of cultural connections and the study of the visual sentiment across languages.

Starting by the fact that either global features and dominant objects bring massive sentiment cues, Sun et al.~\cite{sun2016discovering} proposed an algorithm that extracts and combines features from either the whole image and \virgolette{salient} regions. These regions have been selected by considering proper objectness and sentiment scores  aimed to discover affective local regions.
The proposed method obtained good results compared with~\cite{borth2013large} and~\cite{you2015robust} on three widely used datasets presented in~\cite{borth2013large} and~\cite{you2015robust}.

In~\cite{katsurai2016image} Katsurai and Satoh exploited visual, textual and sentiment features to build a latent embedding space where the correlation between the projected features from different views is maximized. This work implements the CCA (Canonical Correlation Analysis) technique to build a 3-view embedding which provides a tool to encode inputs from different sources (i.e., a text and an image with similar meaning/sentiment are projected nearby in the embedding space) and a method to obtain a sentiment representation of images (by simply projecting an input feature to the latent embedding space). This representation is exploited to train a linear SVM classifier to infer positive or negative polarity. The authors used a composition of RGB histograms, GIST, SIFT based Bag of Words and two mid-level features defined in~\cite{yu2013designing} and~\cite{borth2013large} as visual features. The textual feature is obtained using a Bag of Words approach form text associated to the image, crawled from Flickr and Instagram. The sentiment features are obtained starting from the input text and exploiting an external knowledge base, called SentiWordNet~\cite{esuli2006sentiwordnet}, a well-known lexical resource used in opinion mining to assign sentiment scores to words. 

The works in~\cite{yang2017learning} and in~\cite{ijcai2017-456} perform emotional image classification of images considering multiple emotional labels.
As previously proposed in 2015 by Peng~\cite{peng2015mixed}, instead of training a model to predict only one sentiment label, the authors considered a distribution over a set of pre-defined emotional labels. To this aim, they proposed a multi-task system which optimizes the classification and the distribution prediction simultaneously. 
In~\cite{yang2017learning} the authors proposed two Conditional Probability Neural Networks (CPNN), called Binary CPNN (BCPNN) and Augmented CPNN (ACPNN). 
A CPNN is a neural network with one hidden layer which takes either features and labels as input and outputs the label distribution. Indeed, the aim of a CPNN is to predict the probability distribution over a set of considered labels.
The authors of~\cite{ijcai2017-456} changed the dimension of the last layer of a CNN pre-trained for Object Classification in order to extract a probability distribution with respect to the considered emotional labels, and replaced the original loss layer with a function that integrates the classification loss and sentiment
distribution loss through a weighted combination. Then the modified CNN has been fine-tuned to predict sentiment distributions.
Since the majority of the existing datasets are built to assign a single emotion ground truth to each image, the authors of~\cite{ijcai2017-456} proposed two approaches to convert the single labels to emotional distribution vectors, which elements represent the degree to which each emotion category is associated to the considered image.
This is obtained considering the similarities between the pairwise emotion categories~\cite{plutchik1980general}.
The experimental results show that the approach proposed in~\cite{ijcai2017-456} outperforms eleven baseline Label Distribution Learning (LDL) methods, including BCPNN and ACPNN proposed in~\cite{yang2017learning}.

In~\cite{campos2017pixels} the authors extended their previous work~\cite{campos2015diving} in which they first trained a CNN for sentiment analysis and then empirically studied the contribute of each layer. 
In particular, they used the activations in each layer to train different linear classifiers. In this work the authors also studied the effect of weight initialization for fine-tuning by changing the task (i.e., the output domain) for which the fine-tuned CNN has been originally trained.
Then, the authors propose an improved CNN architecture based on the experimental results and observations. 
Then, the authors propose an improved CNN architecture based on the empirical insights. 


The authors of~\cite{vadicamo2017cross} exploited a large tweets dataset named Twitter for Sentiment Analysis~(T4SA) to finetune existing CNN previously trained for objects and places classification (VGG19~\cite{simonyan2014very} and HybridNet~\cite{zhou2014learning}) for the task of Visual Sentiment Analysis, where the image labels have been defined based on the tweet texts. 
The proposed system has been compared with the CNN and PCNN presented in~\cite{you2015robust}, DeepSentiBank~\cite{chen2014deepsentibank} and MVSO~\cite{jou2015visual} obtaining better results on the built dataset.

\begin{table*}[t]
	\caption{Summary of the most relevant publications on Visual Sentiment Analysis. While the early methods were mostly based on hand crafted visual features, more recent approaches exploit textual metadata and features learned directly from the raw image~(i.e., CNN based representations).}
	\label{tabStateoftheart}
	\centering
	\small
	\scalebox{0.9}{
		\begin{tabular}{|c|c|c|c|}
			\hline
			\textbf{Year} & \textbf{Paper}                                     & \textbf{Input}                                                                 & \textbf{Output}                                                                                                       \\ \hline
			2010          & Siersdorfer et al.~\cite{siersdorfer2010analyzing} & Hand crafted visual features                                                   & Sentiment Polarity                                                                                                    \\ \hline
			2010          & Machajdik et al.~\cite{machajdik2010affective}     & Hand crafted visual features                                                   & Emotional Classification~\cite{mikels2005emotional}                                                                   \\ \hline
			2013          & Borth et al.~\cite{borth2013large}                 & ANP output responses~\cite{borth2013large}                                     & \begin{tabular}[c]{@{}c@{}}Sentiment Polarity and \\ Emotional Classification~\cite{mikels2005emotional}\end{tabular} \\ \hline
			2013          & Yuan et al.~\cite{yuan2013sentribute}              & Hand crafted visual features                                                   & Sentiment Polarity                                                                                                    \\ \hline
			2014          & Yang et al.~\cite{yang2014your}                    & Hand crafted visual features                                                   & Emotional Classification~\cite{mikels2005emotional}                                                                   \\ \hline
			2014          & Chen et al.~\cite{chen2014deepsentibank}           & Raw image                                                                      & ANP annotation~\cite{borth2013large}                                                                                  \\ \hline
			2014          & Xu et al.~\cite{xu2014visual}                      & CNN activations                                                                & 5-scale Sentiment Score                                                                                               \\ \hline
			2015          & You et al.~\cite{you2015robust}                    & Raw image                                                                      & Sentiment Polarity                                                                                                    \\ \hline
			2015          & Peng et al.~\cite{peng2015mixed}                   & Hand crafted visual features                                                   & Distribution of Emotions                                                                                              \\ \hline
			2015          & Wang et al.~\cite{wang2015unsupervised}            & Textual metadata                                                               & Sentiment Polarity                                                                                                    \\ \hline
			2015          & Campos et al.~\cite{campos2015diving}              & Raw image                                                                      & Sentiment Polarity                                                                                                    \\ \hline
			2016          & Sun et al.~\cite{sun2016discovering}               & Image salient regions                                                          & Sentiment Polarity                                                                                                    \\ \hline
			2016          & Katsurai et al.~\cite{katsurai2016image}           & \begin{tabular}[c]{@{}c@{}}Hand crafted visual features\\\& textual metadata\end{tabular} & Sentiment Polarity                                                                                                    \\ \hline
			2017          & Yang et al.~\cite{yang2017learning}                & Raw image                                                                      & Distribution of Emotions                                                                                              \\ \hline
			2017          & Yang et al.~\cite{ijcai2017-456}                   & Raw image                                                                      & Distribution of Emotions                                                                                              \\ \hline
			2017          & Campos et al.~\cite{campos2017pixels}              & Raw image                                                                      & Sentiment Polarity                                                                                                    \\ \hline
			
			2017          & Vadicamo et al.~\cite{vadicamo2017cross}              & Raw image                                                                      & Sentiment Polarity                                                                                                    \\ \hline

			2018          & Li et al.~\cite{li2018image}                       & Text from the inferred ANPs ~\cite{borth2013large}                                                     & Sentiment Polarity                                                                                                     \\ \hline
			
			2018          & Song et al. \cite{song2018boosting}                       & Raw Image                                                      & \begin{tabular}[c]{@{}c@{}}Sentiment Polarity\\\& Emotional Classification~\cite{mikels2005emotional}\end{tabular}                                                                                                    \\ \hline
			
			2018          & Ortis et al. \cite{ortis2018visual}                       & \begin{tabular}[c]{@{}c@{}}Hand crafted visual features\\\& text extracted from the image\end{tabular}                                                      & Sentiment Polarity                                                                                                    \\ \hline
			2019          & Felicetti et al.
			~\cite{felicetti2019visual}                       & \begin{tabular}[c]{@{}c@{}}Raw image\\\& textual metadata\end{tabular}                                                      & Sentiment Polarity                                                                                                    \\ \hline
			
			2019          & Zhu et al. ~\cite{zhu2019joint}                       & \begin{tabular}[c]{@{}c@{}}Raw image\\\& textual metadata\end{tabular}                                                      & Sentiment Polarity                                                                                                    \\ \hline

			2019          & Huang et al. ~\cite{huang2019image}                       & \begin{tabular}[c]{@{}c@{}}Raw image\\\& textual metadata\end{tabular}                                                      & Sentiment Polarity                                                                                                    \\ \hline

			2019          & Corchs et al.
			~\cite{corchs2019ensemble}                       & \begin{tabular}[c]{@{}c@{}}Raw image\\\& textual metadata\end{tabular}                                                      & Emotional Classification~\cite{mikels2005emotional}                                                                                                    \\ \hline

			2019          & Campos et al.
			~\cite{campos2019sentiment}
			& \begin{tabular}[c]{@{}c@{}}Raw image\\\& textual metadata\end{tabular}                                                      & Sentiment Polarity                                                                                                    \\ \hline
			
			2019          & Fortin et al.
			~\cite{fortin2019multimodal}                       & \begin{tabular}[c]{@{}c@{}}Raw image\\\& textual metadata\end{tabular}                                                      & Sentiment Polarity                                                                                                    \\ \hline
			
			2019          & Wu et al.
			~\cite{wu2019visual}                       & \begin{tabular}[c]{@{}c@{}}Raw image\\\& textual metadata\end{tabular}                                                      & Sentiment Polarity                                                                                                    \\ \hline
			
		\end{tabular}
	}

\end{table*}

The system proposed in~\cite{li2018image} represents the sentiment of an image by extracting a set of ANPs describing the image. Then, the weighted sum of the extracted textual sentiment values is computed, by using the related ANP responses as weights. The approach proposed in this paper takes advantage of the sentiment of the text composing the ANPs extracted from images, instead of only considering the ANP responses defined in SentiBank~\cite{borth2013large} as mid-level representations. 
In particular, the sentiment value of an extracted ANP is defined by summing the sentiment scores defined in SentiWordNet~\cite{esuli2006sentiwordnet} and SentiStrength~\cite{thelwall2010sentiment} for the pair of adjective and the noun words of th ANP. 
A logistic regressor is used to infer the sentiment orientation by exploiting the scores extracted from the textual information, and a logistic classifier is trained for polarity prediction by exploiting the traditional ANP responses as representations. Then, the two schemes are combined by employing a late fusion approach.
The authors compared their method with respect to three baselines: a logistic regression model based on the SentiBank mid-level representation, the CNN and PCNN methods proposed in~\cite{you2015robust}.
Experiments show that the proposed late fusion method outperforms the method based only on the mid-level representation defined in SentiBank, demonstrating the contribute given by the sentiment coefficients associated to the text composing the extracted ANPs. However, the CNN and PCNN approaches proposed in~\cite{you2015robust} exhibit better performances than the late fusion method.
The problem of the noise in the text associated to social images by users is very common in the state of the art approaches. 

The work in~\cite{song2018boosting} presents a method for Visual Sentiment Analysis that integrates saliency detection into visual attention learning aimed to let the sentiment classifier focus on salient objects in the image.
In particular, a fully connected neural network is trained to extract salient regions in the input image using multi-scale feature maps. During prediction, the image weighted by the attention map is fed to a CNN trained for Visual Sentiment Analysis. The proposed approach has been evaluated on the tasks of polarity prediction and emotion classification using the Mikels' 8 emotional model~\cite{mikels2005emotional}.

The paper in~\cite{ortis2018visual} addresses the challenge of image sentiment polarity prediction by proposing a novel source of text for this task, dealing with the issue related to the use of text associated to images provided by users, which is commonly used in most of the previous works and is often noisy due its subjective nature. In particular, the authors built an image sentiment classifier that exploits a representation based on both visual and textual features extracted from an image, and evaluated the performances obtained by using the text provided by users (subjective) and the text extracted from the visual content by using four deep models trained on different tasks. Several experiments have been performed by combining different features based on subjective (i.e., user-provided) and objective (i.e., automatically extracted) text. The experiments revealed that employing a source of text automatically extracted from the images, in lieu of the text provided by users, improves the classifier performances.  

In~\cite{felicetti2019visual} the authors combined the visual and textual information of social media posts to predict the overall sentiment of daily news. The system has been evaluated on a dataset of images crawled from Instagram.

The authors of~\cite{zhu2019joint} proposes a joint visual-textual sentiment analysis system trying to exploit more than one modality. However, the authors further incorporated a cross-modality attention mechanism and semantic embedding learning based on bidirectional recurrent neural networks with the aim to design a model able to focus on the visual and textual features that mostly contribute to the sentiment classification. The intuition beyond this approach is that not both text and image contribute equally to the sentiment classification, as visual and textual information differ in their contribution to sentiment analysis. The attention mechanism allows to properly weight input that may deliver inconsistent sentiment, such as the noisy text provided by users~\cite{ortis2018visual}.

Huang et al.~\cite{huang2019image} propose an image-text sentiment model named Deep Multimodal Attentive Fusion (DMAF). This approach defines two attention models aimed to learn effective sentiment classifiers for visual and textual inputs, named Visual Attention Model and Semantic Attention Model respectively. Furthermore, an additional Multimodal Attention Model is employed with the aim to exploit the correlation between visual and textual features. Then, a late fusion approach is used to combine the three attention models.
In~\cite{campos2019sentiment} the authors trained a joint embedding for images and text for sentiment analysis. In particular, they formulated an Adjective Noun Pair~(ANP) detection task in images computed on features extracted from an continuous embedding space that encodes inter-concept relationships between different ANPs. 
In~\cite{fortin2019multimodal} the authors presented a multitask framework to combine multimodal information (i.e., images and text) when it is available, while being able to handle the cases when a modality is missing. In particular, this approach adds two auxiliary image-based and text-based classifiers to the classic multimodal framework. 
Experiments empirically demonstrated that such a multi-task learning approach can also improve generalization by acting as a regularization mechanism.
The work in~\cite{corchs2019ensemble} presents a method that combines visual and textual features by employing an ensemble learning approach. In particular, the authors combined five state of the art classifiers based on visual and textual inputs. The system classifies a givne image using the 8 emotions scheme proposed by Mikels~\cite{mikels2005emotional}.
The authors of~\cite{wu2019visual} combined the use of CNNs and saliency detection to develop a system that first performs a prediction of sentiment of the whole image. Then, if salient regions are detected, performs the sentiment prediction on the sub-images that depict such sub-areas. Finally, the global and local predictions are fused by a weighted sum.

The paper in~\cite{zhao2018affective} presents an overview in the field of affective image content analysis with respect to two main tasks, namely affective gap and perception subjectivity. 

Soleymani et al.~\cite{soleymani2017survey} present a brief overview of the methods adopted to address three different fields: multi-modal spoken reviews and vlogs, sentiment analysis in humans-machines face to face interactions and sentiment of images and tags in social media.

The paper in~\cite{li2019survey} presents an overview on textual sentiment analysis of social networks contents. Then, a focus on multi-modal approaches that combines textual and visual features for visual sentiment analysis is presented. 
\\
The works described in this section have led to significant improvements in the field of Visual Sentiment Analysis.
However these works address the problem considering different emotion models, datasets and evaluation methods (see Table~\ref{tabStateoftheart}).
So far, researchers formulated this task as a classification problem among a number of polarity levels or emotional categories, but the number and the type of the emotional outputs adopted for the classification are arbitrary. The difference in the adopted emotion categories makes result comparison difficult.
Moreover, there is not a strong agreement in the research community about the use of an universal benchmark dataset. Indeed, several works evaluated their methods on their own datasets.
Many of the mentioned works present at least one of the said issues.
\\
Previous surveys give a brief overview of the field followed by a focus on specific issues~\cite{soleymani2017survey,zhao2018affective,li2019survey}. In contrast, the proposed study introduces the basic task in Visual Sentiment Analysis~(i.e., classifying an image as positive or negative) and then extends the discussion by deeply analysing each aspect, covering a larger number of topics, derived tasks and new challenges related to Visual Sentiment Analysis with high level of detail.

\begin{figure}[t]
	\centering
	\includegraphics[width=0.6\linewidth]{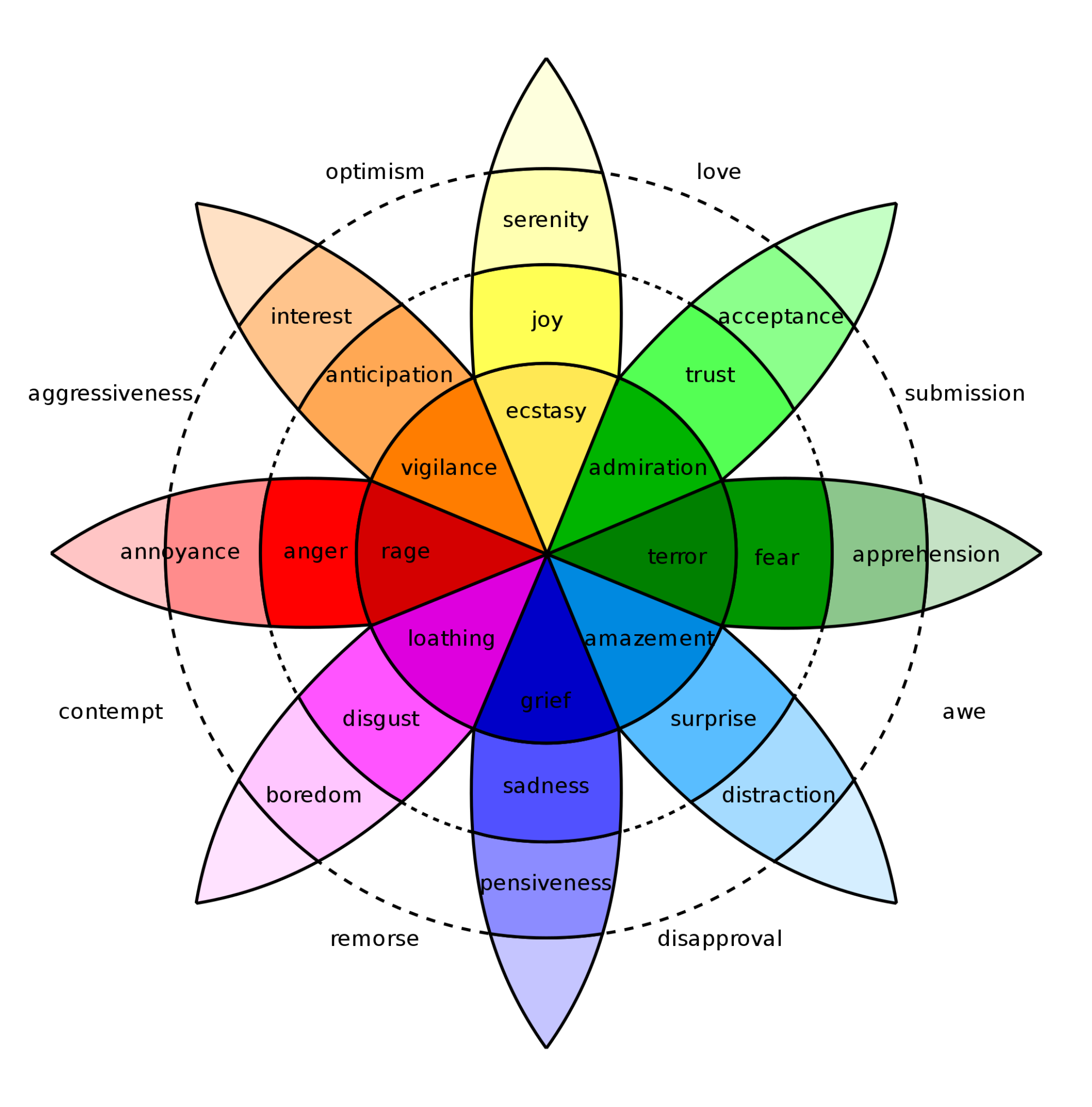}
	\caption{Plutchik's Wheel of Emotions.}
	\label{figWheel}
\end{figure}

\section{Visual Sentiment Analysis Systems}\label{systems}

This section provides an overview of the system design choices, with the aim to provide a comprehensive debate about each specific issue, with proper references to the state of the art.

\begin{figure*}[t]
	\centering
	\begin{minipage}{.35\textwidth}
		\centering
		\includegraphics[width=0.75\linewidth]{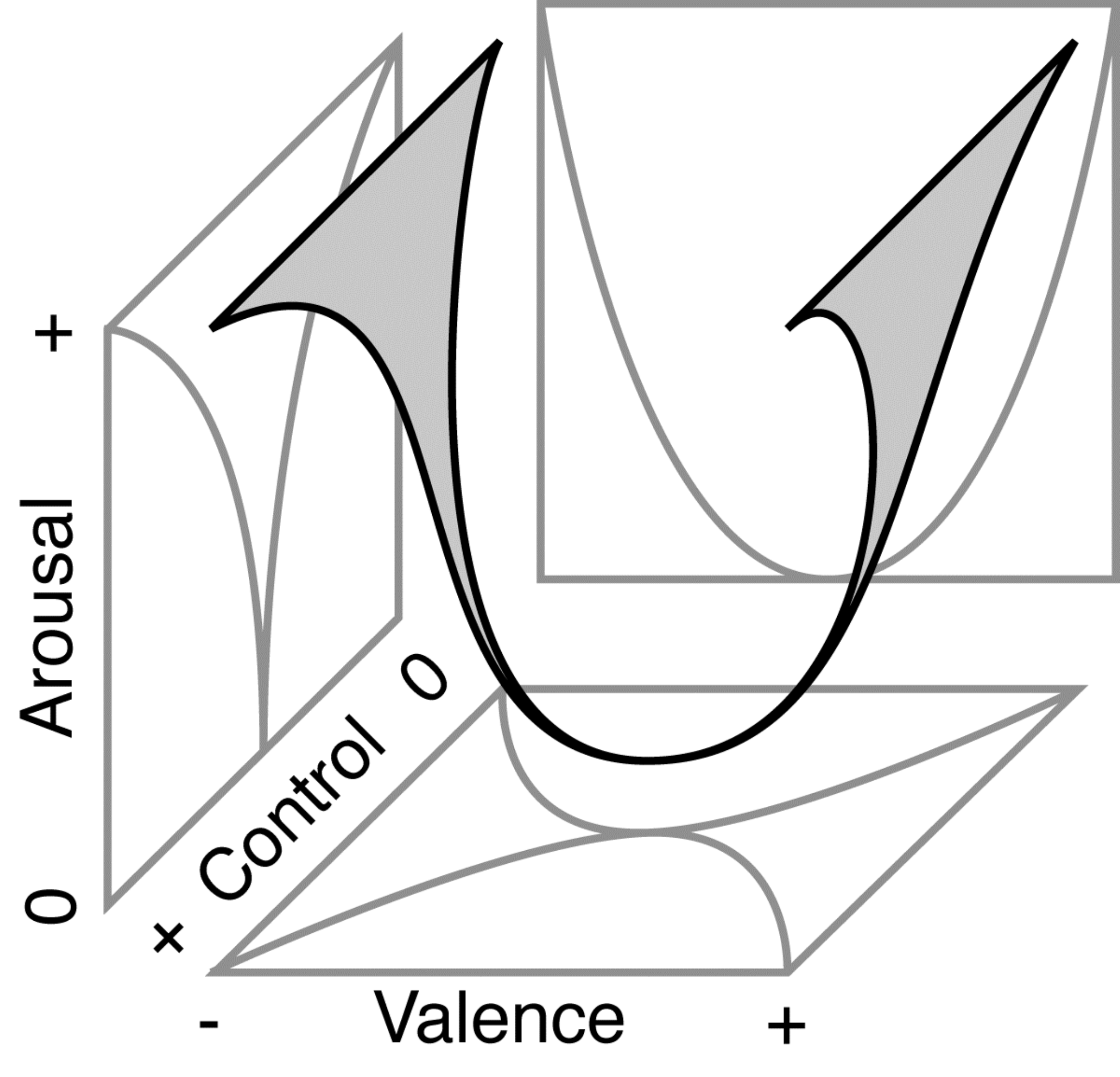}		
		\caption{Illustration of the 3D emotion space (VAC space), image borrowed from~\cite{hanjalic2005affective}
			.}
		\label{figVAC}
	\end{minipage}
	\hspace{0.6cm}
	\begin{minipage}{.35\textwidth}
		\centering
		\includegraphics[width=0.75\linewidth]{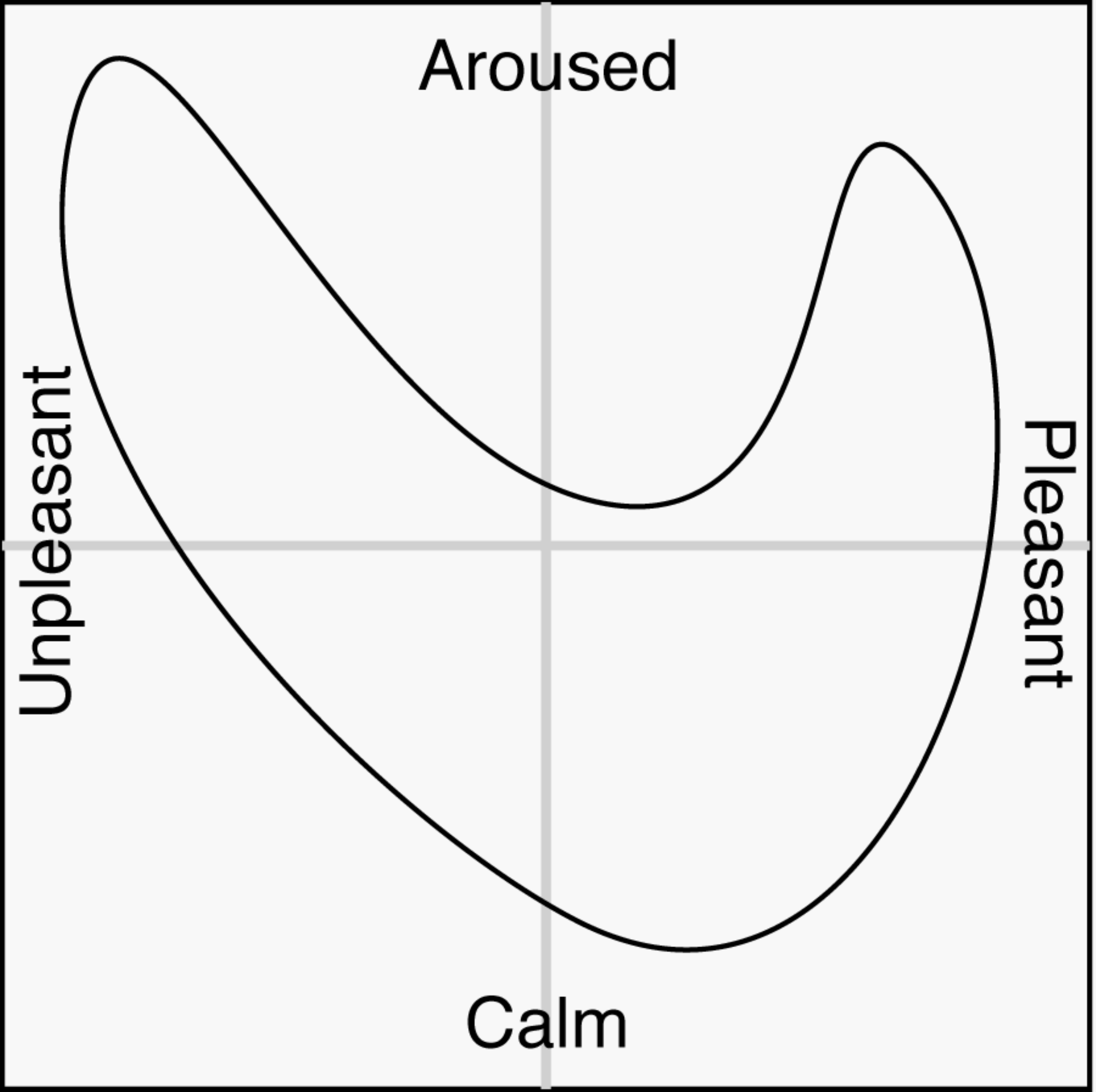}
		\caption{Illustration of the 2D emotion space (VA space) obtained considering just the arousal and the valence axis. Image borrowed from~\cite{hanjalic2005affective}
			.}
		\label{figVA}
	\end{minipage}
	\vspace{-0.5cm}
\end{figure*}

\subsection{How to represent the emotions?}\label{secEmotionModels}
Basically, the goal of a Visual Sentiment Analysis system is to determine the sentiment polarity of an input image (i.e., positive or negative). Several works aims to classify the sentiment conveyed by images into 2 (positive, negative) or 3 polarity levels (positive, neutral, negative)~\cite{borth2013large, siersdorfer2010analyzing,you2015robust}. However, there are also systems that adopt more then 3 levels, such as the 5-level sentiment scheme used by Xu et al.~\cite{xu2014visual} or the 35 ``impression words" used by Hayashi et al.~\cite{hayashi1998image}. 
Beside the polarity estimation, there are systems that perform the sentiment classification by using a set of emotional categories, according to an established emotion model based on previous psychological studies. However, each emotional category usually corresponds to a positive or negative polarity~\cite{machajdik2010affective}. Thus, these systems can be evaluated also for the task of polarity estimation.
Generally, there are two main approaches for emotion modelling:
\begin{itemize}
	\item \textit{\textbf{Dimensional approach:}} emotions are represented as points in a 2 or 3 dimensional space~(see Figures~\ref{figVAC} and~\ref{figVA}). 
	Indeed, as discussed in several studies~\cite{bradley1994emotional, lang1993network, osgood1952nature, russell1977evidence}, emotions have three basic underlying dimensions: valence, arousal and control (or dominance). 
	However, as can be seen from Figure~\ref{figVAC}, the control dimension has a small effect. Therefore, a 2D emotion space is often 
	used, obtained by considering only the arousal and the valence axis. 
	
	\item \textit{\textbf{Category approach:}} according to this model there are a number of basic emotions. Having just a few emotional categories is convenient for tasks such as indexing and classification. This set of descriptive words can be assigned to regions of the VAC (Valence-Arousal-Control) space. Thus it can be considered
	a quantized version of the dimensional approach. 
	The choice of the emotional categories in not an easy task. Since emotion belongs to the psychology domain, the insights and achievements from cognitive science can be beneficial for this problem.
\end{itemize}


What are the basic emotions? There are several works that aim to ask this question. As we observed in Section~\ref{stateoftheart}, the most adopted model is the Plutchnik's Wheel of Emotions~\cite{plutchik1980general}. 
This model defines 8 basic emotions with 3 valences each. Thus it defines a total of 24 emotions~(see Figure~\ref{figWheel}).

According to Ekman's theory~\cite{ekman1987universals} there are just five basic emotions (\virgolette{anger}, \virgolette{fear}, \virgolette{disgust}, \virgolette{surprise} and \virgolette{sadness}).
Another emotions categorization is the one defined in a psychological study by Mikels et al.~\cite{mikels2005emotional}. In this work the authors perform an intensive study on the \textit{International Affective Picture System} (IAPS) in order to extract a categorical structure of such dataset~\cite{lang1999international}. As result, a subset of IAPS have been categorized in eight distinct emotions: \virgolette{amusement}, \virgolette{awe}, \virgolette{anger}, \virgolette{contentment}, \virgolette{disgust}, \virgolette{excitement}, \virgolette{fear} and \virgolette{sad}.
A deeper list of emotion is described in Shaver et al.~\cite{shaver1987emotion}, where emotion concepts are organized in a hierarchical structure. 
Furthermore, a cognitive perspective analysis on basic emotions and their relations is presented in~\cite{izard1992basic}.
\\
\\
As we discussed in this Section, there is a wide range of research on identification of basic emotions. By way of conclusion, the 24 emotions model defined in Plutchnik's theory~\cite{plutchik1980general} is a well established psychological model of emotions. This model is inspired by chromatics in which emotions are organized along a wheel scheme where bipolar elements are placed opposite one to each other. Moreover the three intensities provides a richer set of emotional valences. For these reasons it can be considered the reference model for the identification of the emotional categories.

\subsection{Existing datasets}
There are several sources that can be exploited to build Sentiment Analysis datasets. The general procedure to obtain a human labelled set of data is to perform surveys over a large number of people, but in the context of Sentiment Analysis the collection of huge opinion data can be alternatively obtained by exploiting the most common social platforms (Instagram, Flickr, Twitter, Facebook, etc.), as well as websites for collecting business and products reviews (Amazon, Tripadvisor, Ebay, etc.). Indeed, nowadays people are used to express their opinions and share their daily experiences through the Internet Social Platforms.
\\
\\
In the context of Visual Sentiment Analysis, one of the first published dataset is the International Affective Picture System (IAPS)~\cite{lang1999international}. This dataset has been developed with the aim to produce a set of evocative color images that includes contents from a wide range of semantic categories. This work provides a set of standardized stimuli for the study of human emotional process. 
The dataset is composed by hundreds of pictures related to several scenes including images of insects, children, portraits, poverty, puppies and diseases, which have been manually rated by humans by means of affective words.
This dataset has been used in~\cite{machajdik2010affective} in combination with other two datasets built by the authors, which are publicly available\footnote{www.imageemotion.org}.

In~\cite{yanulevskaya2008emotional} the authors considered a subset of IAPS extended with subject annotations to obtain a training set categorized in distinct emotions according to the emotional model described in~\cite{mikels2005emotional}~(see Section~\ref{secEmotionModels}). 
However, the number of images of this dataset is very low.

In~\cite{machajdik2010affective}, the authors presented the Affective Image Classification Dataset. It consists of two image sets: one containing 228 abstract painting and the other containing 807 artistic photos. These images have been labelled by using the 8 emotions defined in~\cite{mikels2005emotional}.

The authors of the dataset presented in~\cite{siersdorfer2010analyzing} considered the top 1.000 positive and negative words in SentiWordNet~\cite{esuli2006sentiwordnet} as keywords to search and crawl over 586.000 images from Flickr. The list of image URLs as well as the collected including title, image resolution, description and the list of the associated tags is available for comparisons~\footnote{http://www.l3s.de/~minack/flickr-sentiment}.

The Geneva Affective Picture Database (GAPED)~\cite{dan2011geneva} dataset includes 730 pictures labelled considering negative (e.g., images depicting human rights violation scenes), positive (e.g., human and puppies) as well as neutral pictures which show static objects.
All dataset images have been rated considering the valence, arousal, and the coherence of the scene.
The dataset is available for research purposes~\footnote{http://www.affective-sciences.org/home/research/materials-and-online-research/research-material/}.

In 2013 Borth et al.~\cite{borth2013large} proposed a very large dataset ($\sim$~0.5 million) of pictures gathered from social media and labelled with ANP (Adjective Noun Pair) concepts. Furthermore, they proposed Twitter benchmark dataset which includes 603 tweets with photos. It is intended for evaluating the performance of automatic sentiment prediction using features of different modalities (text only, image only, and text-image combined). This dataset has been used by most of the state-of-the-art works as evaluation benchmark for Visual Sentiment Analysis, especially when the designed approaches involve the use of Machine Learning methods such as in~\cite{you2015robust} and in~\cite{yuan2013sentribute} for instance, due to the large scale of this dataset.

Marchewka et al.~\cite{marchewka2014nencki} presented the Nencki Affective Picture System (NAPS), a dataset of high-quality photos depicting people, faces, animals, objects and landscapes. The photos have been manually rated by 204 participants using a 3D dimensional approach and a 9-point rating scale for each dimension. 

The Emotion6 dataset, presented and used in~\cite{peng2015mixed}, has been built considering the Elkman's 6 basic emotion categories~\cite{ekman1987universals}. The number of images is balanced over the considered categories and the emotions associated with each image is expressed as a probability distribution instead of as a single dominant emotion.

%

In~\cite{you2015robust} You et al. proposed a dataset with 1,269 Twitter images labelled into positive or negative by 5 different annotators. Given the subjective nature of sentiment, this dataset has the advantage to be manually labelled by human annotators, differently than other datasets that have been created collecting images by automatic systems based on textual tags or predefined concepts such as the VSO dataset used in~\cite{borth2013large}. 

\begin{table*}[t]
	\small
	\centering
	\caption{Main benchmark datasets for Visual Sentiment Analysis. Some datasets contains several additional information and annotations. All the listed datasets are publicly available.}
	\label{tabDataset}
	\scalebox{0.85}{
		\begin{tabular}{|c|c|c|c|c|c|c|}
			\hline
			\textbf{Year} & \textbf{Dataset}                                                                                                 & \textbf{Size}                                     & \textbf{Labelling}     
			& \begin{tabular}[c]{@{}c@{}}\textbf{Social} \\ \textbf{Media}\end{tabular}                       & \textbf{Polarity}                                   
			& \begin{tabular}[c]{@{}c@{}}\textbf{Additional} \\ \textbf{Metadata}\end{tabular}                                                        
			\\ \hline
			1999          & IAPS~\cite{lang1999international}                                                                              & 716   photos                                            & \begin{tabular}[c]{@{}c@{}}Pleasure, arousal and\\  dominance\end{tabular}                               & \xmark 
			&  \xmark                                              	& \xmark	
			\\ \hline
			2008          &Yanulevskaya et al.~\cite{yanulevskaya2008emotional}                                                                                     & 369 photos                                              & \begin{tabular}[c]{@{}c@{}}Awe, amusement,\\ contentment, excitement,\\disgust, anger, fear, sad\end{tabular}                                   
			& \xmark 
			&  \xmark
			& \xmark
			\\ \hline
			2010          & \begin{tabular}[c]{@{}c@{}}Affective Image\\Classification\\Dataset~\cite{machajdik2010affective}\end{tabular} & \begin{tabular}[c]{@{}c@{}}
				228 paintings\\807 photos
			\end{tabular}                      & \begin{tabular}[c]{@{}c@{}}Awe, amusement,\\ contentment, excitement,\\disgust, anger, fear, sad\end{tabular}                                        
			& \xmark 
			&  \xmark
			& \xmark
			\\ \hline
			2010          & Flickr-sentiment~\cite{siersdorfer2010analyzing}                                                                & 586.000 Flickr photos                             & Positive, negative.                                                                                                 & \checkmark 
			&  \checkmark                                              	& \checkmark                                                   
			\\ \hline
			2011          & GAPED~\cite{dan2011geneva}                                                  & 730 pictures                                      & \begin{tabular}[c]{@{}c@{}}Positive, negative,\\  neutral.\end{tabular}                                                                                                                                         & \xmark 
			&  \checkmark                                              	& \xmark   
			\\ \hline
			2013          & VSO~\cite{borth2013large}                                                                                        &
			\begin{tabular}[c]{@{}c@{}} 0,5 M Flickr Photos\\603 Twitter Images          
			\end{tabular}
			& \begin{tabular}[c]{@{}c@{}}
				- Adjective-Noun Pairs\\
				- Positive or negative
			\end{tabular}    
			
			& \checkmark 
			&  \checkmark                                              	& \checkmark  			
			\\ \hline
			2014          & NAPS~\cite{marchewka2014nencki}                                                                                        &
			\begin{tabular}[c]{@{}c@{}} 1.356 Photos          
			\end{tabular}
			& \begin{tabular}[c]{@{}c@{}}Valence, arousal and\\ approach-avoidance
			\end{tabular}    
			
			& \xmark 
			&  \xmark                                              	& \xmark  			
			\\ \hline
			2015          & Emotion6~\cite{peng2015mixed}                                                                                  & 1.980 Flickr photos                               
			& \begin{tabular}[c]{@{}c@{}}
				- Valence-Arousal score\\
				- 7 emotions distribution 
			\end{tabular}
			& \checkmark 
			&  \xmark                                              	& \xmark  
			\\ \hline
			2015          & You et al.~\cite{you2015robust}                                                                                                   & 1.269 Twitter images                              & Positive, negative.            
			& \checkmark 
			&  \checkmark                                              	& \xmark  
			\\ \hline
			2016          & IESN~\cite{zhao2016predicting}                                                                        & \begin{tabular}[c]{@{}c@{}}
				1M Flickr photos
			\end{tabular} & 
			\begin{tabular}[c]{@{}c@{}}
				- Valence, arousal, dominance\\
				- Awe, amusement,\\ contentment, excitement,\\disgust, anger, fear, sad\\
				- Positive, negative.
			\end{tabular}                                            
			& \checkmark 
			&  \checkmark                                              	& \checkmark  

			\\ \hline
			2016          & CrossSentiment~\cite{katsurai2016image}                                                                         & \begin{tabular}[c]{@{}c@{}}
				90.139 Flickr photos\\65.439 Instagram images
			\end{tabular} & Positive, negative, neutral.                                            
			& \checkmark 
			&  \checkmark                                              	& \xmark  
			\\ \hline
			2017          & T4SA~\cite{vadicamo2017cross}                                                  & 1,5 M Twitter images                          & Positive, negative, neutral.
			& \checkmark 
			&  \checkmark                                              	& \xmark  
			\\ \hline
		\end{tabular}
	}
\end{table*}

The work in~\cite{zhao2016predicting} introduced a large scale dataset named Image-Emotion-Social Net~(IESN), including over 1 million images crawled from Flickr, labelled considering either the dimensional (i.e., valence, arousal and dominance) and the categorical (i.e., Mikels' categories~\cite{mikels2005emotional}) schemes. An important contribute of this work is the inclusion of information related to about 8.000 users and their interactions in the social platform~(e.g., preferences, comments, etc.).
In~\cite{katsurai2016image} the authors crawled two large sets of social pictures from Instagram and Flickr~(CrossSentiment). The list of labelled Instagram and Flickr image URLs is available on the Web~\footnote{http://mm.doshisha.ac.jp/senti/CrossSentiment.html}.


Vadicamo et al.~\cite{vadicamo2017cross} crawled~$\sim3M$ tweets from July to December 2016. The collected tweets have been filtered considering only the ones written in English and including at least an image. 
The sentiment of the text extracted from the tweets has been classified using a polarity classifier based on a paired LSTM-SVM architecture. The data with the most confident prediction have been used to determine the sentiment labels of the images in terms of positive, negative and neutral. The resulting Twitter for Sentiment Analysis dataset~(T4SA) consists of~$\sim1M$ tweets and related~$\sim1.5M$ images.
\\

Datasets such as GAPED, NAPS and IAPS rely on emotion induction. 
This kind of datasets are very difficult to be built in large scale and maintained over time. 
The Machine Learning techniques and the recent Deep Learning methods are able to obtain impressive results as long as these systems are trained with very large scale datasets (e.g., VSO~\cite{borth2013large}). Such datasets can be easily obtained by exploiting the social network platforms by which people share their pictures every day.
These datasets allowed the extensive use of Machine Learning systems that requires large scale datasets. This furthered the building of very large datasets such as T4SA in the last few years.
Table~\ref{tabDataset} summarizes the main dataset just reported with details about the number of images, the source (e.g., Social Platform, paintings, etc.) and the labelling options.

\subsection{Features}
One of the most difficult step for the design of a Visual Sentiment Analysis system, and in general for the design of a data analysis approach is the selection of the data features that better encode the information that the system is aimed to infer.
Image features for Visual Sentiment Analysis can be categorized within three levels of semantics:
\begin{itemize}
	\item \textit{\textbf{Low-level features - }} These features describe distinct visual phenomena in an image mainly related in some way to the color values of the image pixels. They usually includes generic features such as color histograms, HOG, GIST. In the context of Visual Sentiment Analysis, previous works can be exploited to extract particular low-level features derived from proper studies on art and perception theory. These studies suggest that some low-level features, such as colors and texture can be used to express the emotional effect of an image~\cite{machajdik2010affective}.
	\item \textit{\textbf{Mid-level features - }}This group of features bring more semantic, thus they are more interpretable and have stronger associations with emotions~\cite{zhao2014exploring}. One example is given by the scene-based 102-dimensional feature defined in~\cite{yuan2013sentribute}. Furthermore, many of the aforementioned works on Visual Sentiment Analysis exploit the 1200-dimensional mid-level representation given by the 1200 Adjective-Noun Pairs (ANP) classifiers defined by Borth et al.~\cite{borth2013large}.
	\item \textit{\textbf{High-level features - }} These features describe the semantic concepts shown in the images. Such a feature representation can be obtained by using pre-trained classification methods or semantic embeddings~\cite{katsurai2016image}.
\end{itemize}

In 2010 Machajdik and Hanbury~\cite{machajdik2010affective} performed an intensive study on image emotion classification by properly combining the use of several low and high visual features. These features have been obtained by exploiting concepts from and art theory~\cite{itten1973art, valdez1994effects}, or exploited in image retrieval~\cite{stottinger2009translating} and image classification~\cite{datta2006studying, wei2006image} tasks.
They selected 17 visual features, categorized in 4 groups:
\begin{itemize}
	\item \textbf{color:} mean saturation and brightness, 3-dimensional emotion representation by Valdez et al.~\cite{valdez1994effects}, hue statistics, colorfulness measure according to~\cite{datta2006studying}, number of pixels of each of the 11 basic colors~\cite{van2007learning}, Itten contrast~\cite{itten1973art}, color histogram designed by Wang Wei-ning et al.~\cite{wei2006image};
	
	\item \textbf{texture:} wavelet textures for each HSB channel, features by Tamura et al.~\cite{tamura1978textural}, and features based on GLCM (i.e., correlation, contrast, homogeneity, and energy for the HSB channels);
	
	\item \textbf{composition:} the number of resulting segments obtained after the application of a waterfall segmentation (denoted as \virgolette{level of detail} in~\cite{machajdik2010affective}), depth of field (DOF)~\cite{datta2006studying}, statistics on the line slopes by using the Hough transform (denoted as \virgolette{dynamics}), rule of thirds;
	
	\item \textbf{content:} number of detected front faces, number of the biggest face pixels, count of skin pixels, ratio of the skin pixels over the face size.
	
\end{itemize}

Most of the mentioned works in Visual Sentiment Analysis combine huge number of hand-crafted visual features. Although all the exploited features have been proven to have a direct influence on the perceived emotion by previous studies, there is not agreement about which of them give the most of the contribution on the aimed task.
Besides the selection of proper hand-crafted features, designed with the aim to encode the sentiment content conveyed by images, there are other kind of approaches that lean on representation learning techniques based on Deep Learning~\cite{chen2014deepsentibank, xu2014visual, you2015robust}. By employing such representation methods image features are learned from the data. This avoid the designing of a proper feature for the task of Visual Sentiment Analysis, because the system automatically learns how to extract the needed information from the input data. These methods requires huge amounts of labelled training data, and an intensive learning phase, but obtain better performances in general.

Another approach, borrowed from the image retrieval methods, consists on combining textual and visual information through multimodal embedding systems~\cite{katsurai2016image,ortis2018visual,campos2019sentiment,fortin2019multimodal}. In this case, features taken from different modalities (e.g., visual, textual, etc.) are combined to create a common vector space in which the correlations between projections of the different modalities are maximized (i.e., an embedding space).
\\
\begin{figure*}[t]
	\centering
	\includegraphics[width=\linewidth]{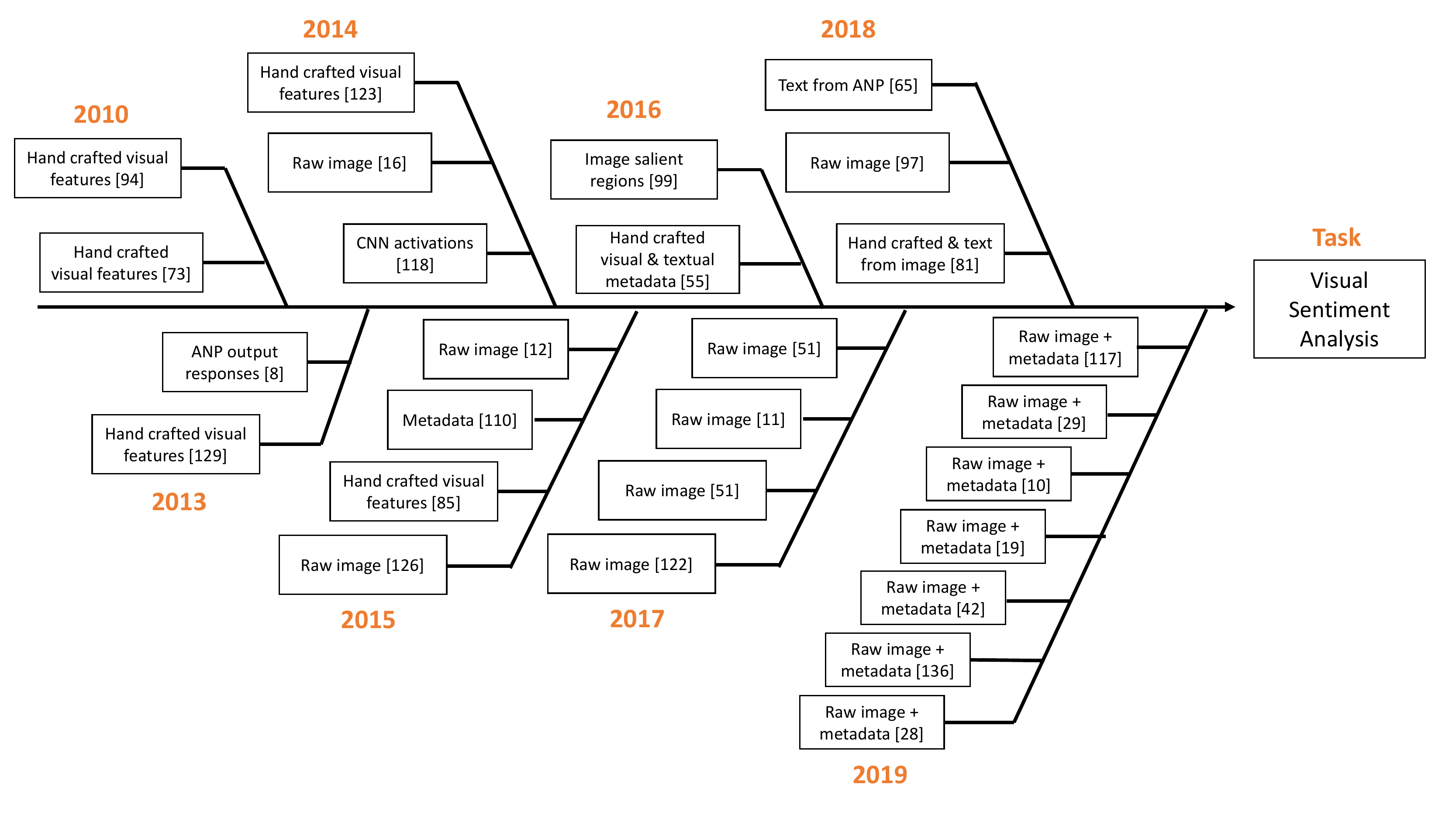}
	\caption{Temporal evolution of the inputs considered by the state-of-the-art works in Visual Sentiment Analysis.}
	\label{figFishbone}
\end{figure*}
\\
In the diagram depicted in Figure~\ref{figFishbone} the related works described in Section~\ref{stateoftheart} are arranged temporally. For each work, the diagram reports the input features considered by the proposed system. It is possible to notice that early works were mostly based on hand-crafted visual features until 2016. Then, most of the works implemented systems able to automatically learn meaningful features from the input raw image~(see branch 2017 in Figure~\ref{figFishbone}). Lastly, the mentioned works published in 2019 combine the contribution of both the raw image and the associated metadata, as they can exploit large dataset built considering images crawled from social media platforms, which offer several additional textual metadata associated to images. 

So far, there is not an established strategy to select the visual features that allows to  address the problem. Most of the previous exploited features demonstrated to be useful, but recent results on Visual Sentiment Analysis 
suggest that it's worth investigating the use of representation learning approaches such as Convolutional Neural Networks and multimodal embeddings that are able to embody the contribute of multiple sources of information.

\section{Problem analysis}\label{problem}


In this section we propose a formulation of the problem, which highlights the related issues and the key tasks of Visual Sentiment Analysis. This allows to better focus the related sub-issues which form the Visual Sentiment Analysis problem and support the designing of more robust approaches. Moreover, to address the overall structure of the problem is useful to suggest a common framework helping researchers to design more robust approaches.
Starting from the definition of the Sentiment Analysis problem applied to the natural language text given by Liu~\cite{liu2012sentiment}, we propose to generalize the definition in the context of Visual Sentiment Analysis.  

Text based Sentiment Analysis can be performed considering different levels of detail: 

\begin{itemize}
	\item at the \textbf{document level} the task is to classify whether a whole document (i.e., the whole input) expresses a positive or negative sentiment. This model works on the underling assumption that the whole input discusses only one topic;
	\item at the \textbf{sentence level} the task is to find each phrase within the input document and determine if each sentence express a positive or negative (or neutral) sentiment;
	\item the \textbf{entity and aspect level} performs finer-grained analysis by considering all the opinions expressed in the input document and defining a sentiment score (positive or negative) for each detected target.
\end{itemize}
Similarly, if the subject of the analysis is an image, we can:
\begin{itemize}
	\item consider a Sentiment Analysis evaluation for the whole image. These systems work with global image features (e.g., color histograms, saturation, brightness, colorfulness, color harmony, etc.);
	\item consider an image as a composition of several sub-images according to its specific content. A number of sub-images is extracted and the sentiment analysis is performed on each sub-image obtained by exploiting methods such as multi-object detection, image segmentation, objectness extraction~\cite{alexe2012measuring};
	\item define a set of image aspects, in terms of low level features, each one associated to a sentiment polarity based on previous studies~\cite{machajdik2010affective,yanulevskaya2008emotional}. 
	This is essentially the most fine-grained analysis to be considered.
\end{itemize}

When a system aims to perform Sentiment Analysis on some textual content, basically it is looking for the opinions in the content and extracting the associated sentiment. An opinion consists of two main components: a target (or topic), and a sentiment. The opinions can be taken from more than one person, this means that the system has to take into account also the opinion holder. Furthermore, opinions can change over time, thus also the time an opinion is expressed has to be taken into account.
According to Liu~\cite{liu2012sentiment}, an opinion (or sentiment) is a quintuple 
\begin{equation}
\left( e_i, a_{ij}, s_{ijkh}, h_k, t_l \right) 
\end{equation}
where \boldmath{$e_i$} is the name of an entity, $a_{ij}$ is an aspect related to $e_i$, $s_{ijkh}$ is the sentiment score with respect to the aspect $a_{ij}$, $h_k$ is the opinion holder, and $t_l$ is the time when the opinion is expressed by the opinion holder $h_k$.
The sentiment score $s_{ijkh}$ can be expressed in terms of polarity, considering positive, negative or neutral polarity; or with different levels of intensity. 
The special aspect \virgolette{GENERAL} is used when the sentiment is expressed for the whole entity. In this case, either the entity $e_{i}$ and the aspect $a_{ij}$ represent the opinion target. 

%

This definition is given in the context of opinion analysis applied on textual contents which express positive or negative sentiments.
In the case of Sentiment Analysis applied on visual contents there are some differences. Indeed, when the input is a text, Sentiment Analysis can easily lean on context and semantic information extracted directly from the text. Thus the problem is to be considered into the NLP domain. 
When the input is an image, because of the \textit{affective gap} between visual content representations and semantic concepts such as human sentiments, the task to associate the visual features with sentiment labels or polarity scores results challenging.
Such \textit{affective gap} can be defined as: 
\\

\textit{\virgolette{the lack of coincidence between the measurable signal properties, commonly referred to as features, and the expected affective state in which the user is brought by perceiving the signal}}~\cite{hanjalic2006extracting}.
\\

In the following paragraphs each of the sentiment components previously defined (entity, aspect, holder and time) are discussed in the context of Visual Sentiment Analysis.

\subsection{Entity and Aspects}
The entity is the subject (or target) of the analysis. In the case of Visual Sentiment Analysis the entity is the input image. 
In general, an entity can be viewed as a set of \virgolette{parts} and \virgolette{attributes}. The set of the entity's parts, its attributes, plus the special aspect \virgolette{GENERAL} forms the set of the aspects. 
This structure can be transferred to the visual domain considering different levels of visual features.
Indeed, as mentioned above, in the case of visual contents, Sentiment Analysis can be performed considering different level of visual detail. The most general approach performs Sentiment Analysis considering the whole image, this corresponds to apply a Visual Sentiment Analysis method on the \virgolette{GENERAL} aspect.
The parts of an image can be defined by considering a set of sub-images. This set can be obtained by exploiting several Computer Vision techniques, such as background/foreground extraction, image segmentation, multi object recognition or dense captioning~\cite{Karpathy_2015_CVPR, karpathy2015deep}. The attributes of an image regards its aesthetic quality features, often obtained by extracting low-level features.
Exploiting this structured image hierarchy, a sentiment score can be achieved for each aspect. Finally, the scores are combined to obtain the sentiment classification~(e.g., data can be used as input features of a regression model).

Instead of representing the image parts as a set of sub-images, an alternative approach can rely on a textual description of the depicted scene. The description of a photo can be focused on a specific task of image understanding. By changing the task, we can obtain different descriptions of the same image from different points of view. Then, these complementary concepts can be combined to obtain the above described structure.
Most of the existing works in sentiment analysis of social media exploit textual information manually associated to images by performing textual Sentiment Analysis. 
Although the text associated to social images is widely exploited in the state-of-the-art to improve the semantics inferred from images, it can be a very noisy source because it is provided by the users; the reliability of such input is often based on the capability and the intent of the users to provide textual data that are coherent with respect to the visual content of the image. There is no guarantee that the subjective text accompanying an image is useful.  
In addition, the tags associated to social images are often selected by users with the purpose to maximize the retrieval and/or the visibility of such images by the platform search engine. In Flickr, for instance, a good selection of tags helps to augment the number of views of an image, hence its popularity in the social platform.
These information are hence not always useful for sentiment analysis. 
Starting from this observation, the authors of~\cite{ortis2018visual} presented a work on Image Polarity Prediction exploiting Objective Text extracted directly from images, and experimentally compared such text with respect to the Subjective~(i.e., user provided) text information commonly used in the state-of-the-art approaches. 
For a deeper analysis, a comprehensive treatise of image tag assignment is presented in~\cite{li2016socializing}.

As discussed in~\cite{gong2014multi}, the semantic of an image can be expressed by means of an object category (i.e., a class). However the tags provided by users usually include several additional terms, related to the object class, coming from a larger vocabulary. As an alternative, the semantic could be expressed by using multiple keywords corresponding to scenes, object categories, or attributes.

\begin{figure*}[t]
	\centering
	\includegraphics[width=1\linewidth]{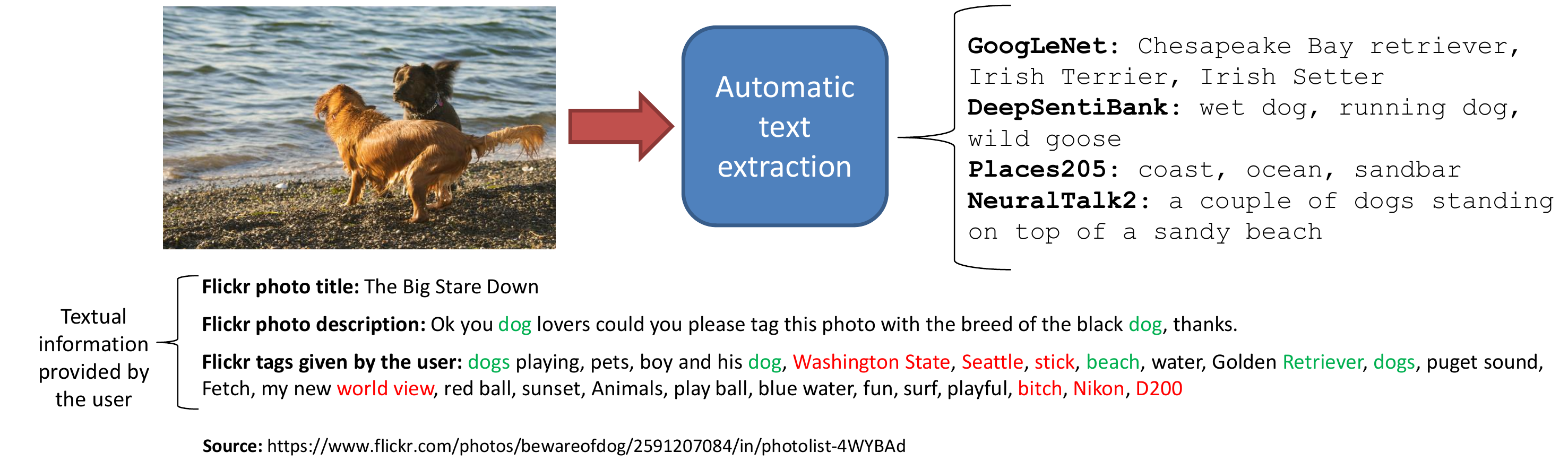}
	\caption{The paper in~\cite{ortis2018visual} extracts the text describing a given image by exploiting four different deep learning architectures. The considered architectures are used to extract text related to objects, scene and image description. The figure shows also the text associated to the image by the user (i.e., title, description and tags) in the bottom. The subjective text presents very noisy words which are highlighted in red. The words that appears in both sources of text are highlighted in green.}
	\label{figFeaturesExtraction}
\end{figure*}

Figure~\ref{figFeaturesExtraction} shows an example taken from~\cite{ortis2018visual}. The textual information below the image is the text provided by the Flickr's user. Namely the photo title, the description and the tags are usually the text that can be exploited to make inferences on the image.
This example shows how the text can be very noisy with respect to any task aimed to understand the sentiment that can be evoked by the picture. Indeed the title is used to describe the tension between the depicted dogs, whereas the photo description is used to ask a question to the community. Furthermore, most of the provided tags include misleading text such as geographical information (i.e., Washington State, Seattle), information related to the camera (i.e., Nikon, D200), objects that are not present in the picture (i.e., boy, red ball, stick) or personal considerations of the user (i.e., my new word view). Moreover, in the subjective text there are many redundant terms (e.g., dog).
Another drawback of the text associated to social images is that two users can provide rather different information about the same picture, either in quality and in quantity. Finally, there is not guarantee that such text is present; this is an intrinsic limit of all Visual Sentiment Analysis approaches exploiting subjective text.

Starting from the aforementioned observations about the user provided text associated to social images, the paper in~\cite{ortis2018visual} exploits an objective aspect of the textual source that comes directly from the understanding of the visual content. This text is achieved by employing a set of deep learning models trained to accomplish different visual inference tasks on the input image. At the top right part of Figure~\ref{figFeaturesExtraction} the text automatically extracted with different scene understanding methods is shown. In this case, the inferred text is very descriptive and each model provides distinctive information related to objects, scene, context, etc.
The objective text extracted by the three different scene understanding methods has a pre-defined structure, therefore all the images have the same quantity of textual objective information. 
Such approach provides an alternative user-independent source of text which describes the semantic of images, useful to address the issues related to the inherent subjectivity of the text associated to images.

Several papers faced the issues related to the subjective text associated to images, such as tag refinement and completion~\cite{li2016socializing,sang2012user,xu2009tag,wu2013tag}, which aims at alleviating the number of noisy tags and enhancing the number of informative tags by modelling the relationship between visual content and tags.

\subsection{Holder}
Emotions are subjective, they are affected by several factors such as gender, individual background, age, environments, etc. However, emotions also have the property of stability~\cite{wang2008survey}. This means that the average emotion response of a statistically large set of observers is stable and reproducible. The stability of emotion response enables researchers to generalize their results, when obtained on large datasets.

Almost all the works in Visual Sentiment Analysis ignore the sentiment holder, or implicitly consider only the sentiment of the image publisher. In this context at least two holders can be defined: the image owner and the image viewer. Considering the example of an advertising campaign, where the owner is the advertising company and the viewers are the potential customers, it's crucial the study and analysis the connection between the sentiment intended by the owner and the actual sentiment induced to the viewers. 

These days, the social media platforms provide a very powerful mean to retrieve real-time and large scale information of people reactions toward topics, events and advertising campaigns.  
The work in~\cite{chen2014predicting} is the first that distinguishes publisher affect (i.e., intent) and viewer affect (i.e., reaction) related to the visual content. 
This branch of research can be useful to understand the relation between the affect concepts of the image owner and the evoked viewer ones, allowing new user centric applications.

Zhao et al.~\cite{zhao2016predicting} proposed a method to predict the personalized emotion induced by an image for each individual viewer (i.e., user in the social platform). To this aim, the authors considered several contextual factors such as the social context, temporal evolution and the location. This work required the definition of an ad-hoc dataset which includes information about a large scale of involved users, beside the set of images.
User profiling helps personalization, which is very important in the field of recommendation systems. The insights that could be obtained from such a research branch can be useful for several business fields, such as advertisement and User Interface design (UI). And the community of user interface designers started to take into account the emotional effect of the user interfaces toward users who are interacting with a website, product or brand. The work in~\cite{lockner2014emotion} discusses about methods to measure user’s emotion during an interface interaction experience, with the aim to assess the interface design emotional effect. 
Progresses in this field promote the definition of new design approaches such as Emotional UI~\cite{emoji2015w}, aimed to exploit the emotions conveyed by visual contents. Indeed, emotions have been traditionally considered to be something that the design evoked, now they represent something that drives the design process. While so far, designers focused on \virgolette{user friendly} design (i.e., interfaces easy to use), now they need to focus on design that stimulates and connects the product with users deeply.
%
%
%
\\
\\
Although the interesting cues discussed in this paragraph, currently the development of Visual Sentiment Analysis algorithms that concern the sentiment holder find difficulties due the lack of specific datasets. In this context, the huge data shared on the social media platforms can be exploited to better understand the relationships between the sentiment of the two main holders (i.e., owner/publisher and viewer/user) through their interactions.


\subsection{Time}
\begin{figure}[t]
	\centering
	\includegraphics[width=0.6\linewidth]{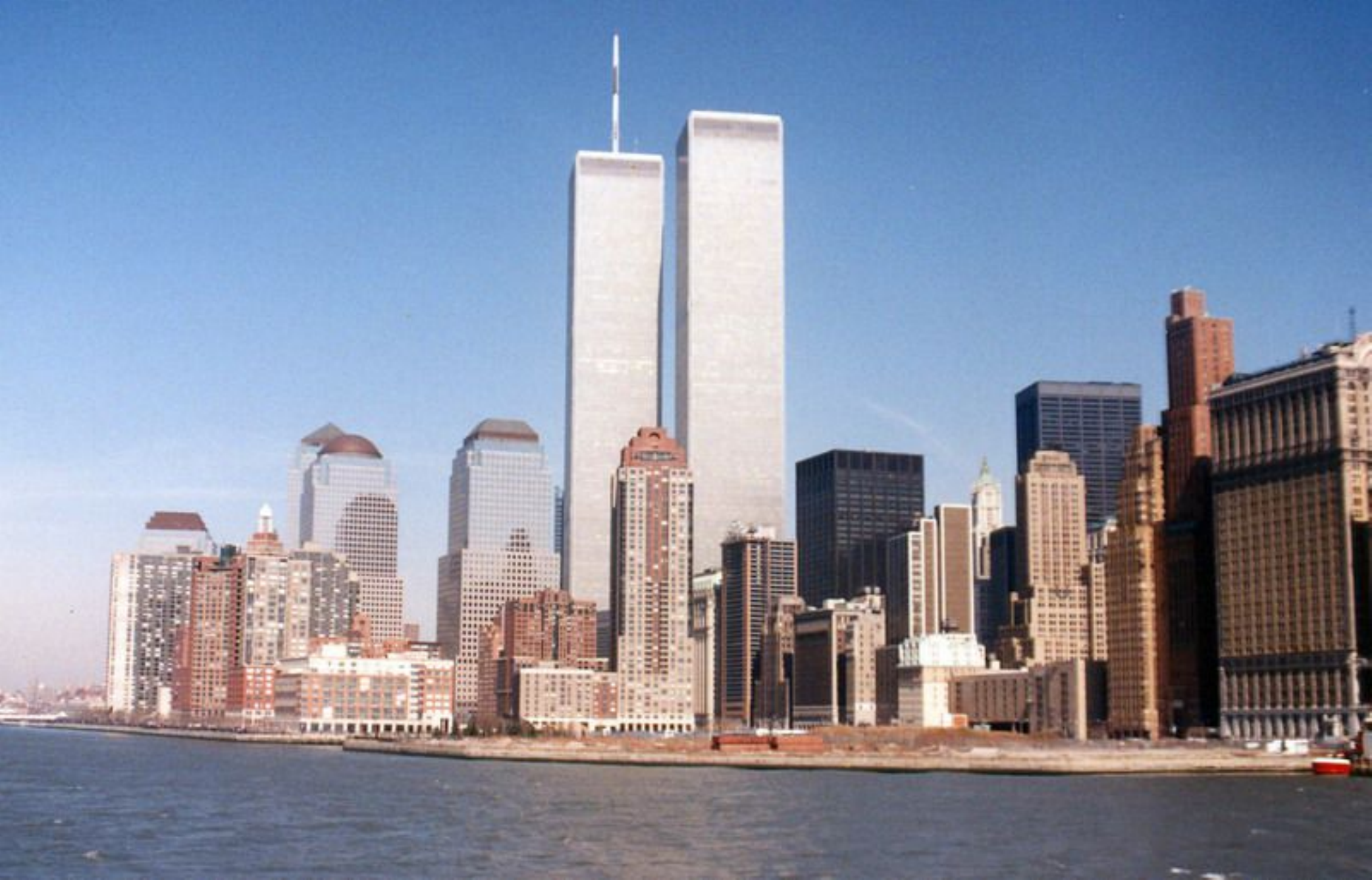}
	\caption{Picture of the World Trade Center, taken in 1990.}
	\label{figTwintowers}
	\vspace{-0.5cm}
\end{figure}
Although almost all the aforementioned works ignore this aspect, the emotion evoked by an image can change depending on the time. 
This sentiment component can be ignored the most of times, but in specific cases is determinant. Moreover, is very difficult to collect a dataset related to the changes in the emotion evoked by images over time.
For example, the sentiment evoked by an image depicting the World Trade Center (Figure~\ref{figTwintowers}) is presumably different if the image is shown before or after 9/11.

Although there are not works on Visual Sentiment Analysis that analyse the changes of image sentiments over time, due to the specificity of the task and the lack of image datasets, there are several works that exploits the analysis of images over time focused on specific cognitive and psychology applications. 
As an example, the work in~\cite{reece2016instagram} employed a statistical framework to detect depression by analysing the sequence of photos posted on Instagram. The findings of this paper suggest the idea that variations in individual psychology reflect in the use social media by the users, hence they can be computationally detected by the analysis of the user's posting history.

In~\cite{khosla2012memorability} the authors studied which objects and regions of an image are positively or negatively correlated with memorability, allowing to create memorability maps for each image. This work provides a method to estimate the memorability of images from many different classes. To collect human memory scores, the adopted experimental procedure consists of showing several occurrences of the same images at variable time intervals. The employed image dataset has been created by sampling images from a number of existing dataset, including images evoking emotions~\cite{machajdik2010affective}.

\section{Challenges}\label{challenges}
So far we discussed on the current state of the art in Visual Sentient Analysis, describing the related issues, as well as the different employed approaches and features.
This section aims to introduce some additional challenges and techniques that can be investigated.

\subsection{Popularity}
One of the most common application field of Visual Sentiment Analysis is related to social marketing campaigns. In the context of social media communication, several companies are interested to analyse the level of people engagement with respect to social posts related to their products. This can be measured as the number of post's views, likes, shares or by the analysis of the comments. These information can be further combined with web search engine and companies website visits statistics, to find correlations between social advertising campaigns and their aimed outcomes (e.g., brand reputation, website/store visits, product dissemination and sale, etc.)~\cite{hanna2011we,goeldi2011website,trusov2009effects}.

The popularity of an image is a difficult quantity to define, hence to measure or infer. 
However human beings are able to predict what visual contents other people will like in specific contexts (e.g., marketing campaigns, professional photography). This suggest that there are some common appealing factors in images.
So far, researches have been trying to gain insights into what features make an image popular.

As mentioned in the previous sections, the quality of the text associated to images is often pre-processed in order to avoid noisy text. 
On the other hand, the users who want increase the reach of their published contents are used to associate popular tags to their images, regardless their relevance with the image content. This is motivated by the fact that image associated tags are used by the image search engines in these platforms, so the use of popular tags highs the reach of the pictures.
Therefore, in this context, the tags associated to images become a crucial factor to understand the popularity of images in social platforms.
For instance, Yamasaki et al.~\cite{yamasaki2014social} proposed an algorithm to estimate the social popularity of images uploaded on Flickr by using only text tags.
Other features should be also taken into account such as: number of user followers and groups, which represent the reach capability of the user. These factors make the task of popularity prediction very different from the task of sentiment polarity classification in the selection of features, methods and measures of evaluation.

The work in~\cite{mcparlane2014nobody} considers the effect of 16 features in the prediction of the image popularity. Specifically, they considered image context (i.e., day, time, season and acquisition settings), image content (i.e., image content provided by detectors of scenes, faces and dominant colors), user context and text features (i.e., image tags).
The authors cast the problem as a binary classification by splitting the dataset between images with high and low popularity measure.
As popularity measures the authors considered the views and the comments counts. Their study highlights that comments are more predictable than views, hence comments are more correlated with the studied features. 
The experimental results show that the accuracy values achieved only considering textual features (i.e., tags) outperform the performances of the classification based on other features and the combinations of them, for both comments and views counts classifications.

In 2014 Khosla et al.~\cite{khosla2014makes} proposed a log-normalized popularity score that has been then commonly used in the community. Let $c_i$ be a measure of the engagement achieved by a social media item (e.g., number of likes, number of views, number of shares, etc.), also known as popularity measure. The popularity score of the $i^{th}$ item is defined as follows:

\begin{equation}\label{eqPopScore}
score_i = \log \left ( \frac{c_i}{T_i} + 1 \right )
\end{equation}
where $T_i$ is the number of days since the uploading of the image on the Social Platform. 
Equation~\ref{eqPopScore} normalizes the number of interactions reached by an image by dividing the engagement measure by the time.
However, the measures $c_i$ related to social posts are cumulative values as they continuously collect the interactions between users and the social posts during their time on-line. Therefore, this normalization will penalize social media contents published in the past with respect to more recent contents, especially when the difference between the dates of posting is high. Indeed, the most of the engagement obtained by a social media item is achieved in the first period, then the engagement measures become more stable. There are very few works which takes into account the evolution of the image popularity over time. For example, the study presented in~\cite{valafar2009beyond} shows that photos obtain most of their engagement within the first 7 days since the date of upload. However, this study is focused on Flickr, and each social platform has its own mechanisms to show contents to users.

In~\cite{khosla2014makes} the authors analysed the importance of several image and social cues that lead to high or low values of popularity. In particular they considered the relevance of user context features (e.g., mean views, number of photos, number of contacts, number of groups, average groups' members, etc.), image context features (e.g., title length, description length, number of tags), as well as image features (e.g., GIST, LBP, BoW color patches, CNN activations features, etc.). Is interesting to notice that, differently than several works on sentiment polarity prediction in which the text concerning the images (i.e., title, description and tags included in the post) is semantically analysed in order to achieve sentiment related insights on the image content, in this work only the length of the text associated to the images is considered. 

Cappallo et al.~\cite{cappallo2015latent} address the popularity prediction problem as a ranking task by exploiting a latent-SVM objective function defined such that the ranking of the popularity scores between pairs of images is maintained.
They considered the number of views and comments for Flickr images and the number of re-tweets and favorites for Twitter.

The problem of image popularity prediction is also addressed in~\cite{gelli2015image}, whose experiments suggest that some sentiment ANPs defined in VSO~\cite{borth2013large} have a correlation with popularity.

In~\cite{almgren2016predicting} the authors considered the number of likes achieved within the first hour after the image posting (early popularity) to predict the popularity after a day, a week or a month. This study has been performed on Instagram images and the dataset is publicly available~\footnote{http://www1bpt.bridgeport.edu/$\sim$jelee/sna/pred.html.}. The images are categorized as popular or not popular considering a popularity threshold obtained with the Pareto principle (80~\% - 20~\%). Three features representing information that is retrieved within the first hour of image upload (i.e., early information) are evaluated: social context (based on the user's number of followers), image semantics (based on image caption and NLP), and early popularity in the first hour. The binary classification is performed by using a Gaussian Naive Bayes Model. The experimental results show that the early popularity feature significantly outperforms the other evaluated features. Furthermore, the authors compared the proposed semantic feature with the features proposed by~\cite{mcparlane2014nobody} for the task of popularity binary classification considering the MIR-1M Flickr dataset~\cite{huiskes2010new}, obtaining better accuracy rates.

Most of the works addressing the problem of popularity prediction follow a very similar pipeline. First, a set of interesting features that have been demonstrating correlation with the images sentiment or popularity is selected. Then a model for each distinctive feature is trained to understand the predictive capability of each feature. In the above discussed works, the popularity prediction task is cast as a ranking or a regression problem. Therefore, the exploited algorithms are ranking SVM and latent SVM in the case of ranking, and SVR for regression.  Then, the features are combined with the aim to improve the performances of the method. The evaluation is usually quantified through the Spearman's correlation coefficient.
\\
\\
Although the task of image popularity prediction is rather new, there are interesting datasets available for the development of massive learning systems (i.e., deep neural networks).
The Micro-Blog Images 1 Million (MBI-1M) dataset is a collection of 1M images from Twitter, along with accompanying tweets and metadata. 
The dataset was introduced by the work in~\cite{cappallo2015latent}. A subset of the the Trec 2013 micro-blog track tweets collection~\cite{lin2013overview} has been selected.

The MIR-1M dataset~\cite{huiskes2010new} is a collection of 1M photos from Flickr. These images have been selected considering the interestingness score used by Flickr to rank images.

The Social Media Prediction~(SMP) dataset is a large-scale collection of social posts, recently collected for the ACM Multimedia 2017 SMP Challenge~\footnote{\small Challenge webpage: https://social-media-prediction.github.io/MM17PredictionChallenge}. This dataset consists of over 850K posts and 80K users, including photos from VSO~\cite{borth2013large} as well as photos collected from personal users' albums~\cite{Wu2016TimeMatters,Wu2017DTCN,Wu2016TemporalPrediction}. 
In particular, the authors aimed to record the dynamic variance of social media data. Indeed, the social media posts in the dataset are obtained with temporal information (i.e., posts sequentiality) to preserve the continuity of post sequences.
Two challenges have been proposed:
\begin{itemize}
	\item \textbf{Popularity Prediction:} the task is to predict a popularity measure defined for the specific social platform (e.g., number of photo's views on Flickr) of a given image posted by a specific user;
	\item \textbf{Tomorrow's Top Prediction:} given a set of photos and the data related to the past photo sharing history, the task is to predict the top-n popular posts (i.e., ranking problem over a set of social posts) on the social media platform in the next day.
\end{itemize}
The SMP dataset includes features such as unique picture id (pid) and associated user id (uid). From these information one can extract almost all the user and photo related data available in Flickr. Some metadata of the picture and user-centered information are also included in the dataset. Moreover, the popularity scores (as defined in Equation~\ref{eqPopScore}) are provided. 
The SMP dataset furthered the development of time aware popularity prediction methods, which exploit time information to define new image representation spaces used to infer the image popularity score at a precise time or at pre-defined time scales. 
Li et al.~\cite{li2017hybrid} extracted multiple time-scale features from a set of timestamps related to the photo post. As instance, the timestamp \virgolette{postdate} is used to define several features with different time scales: \virgolette{season of year}, \virgolette{month of year}, etc.
The framework presented in~\cite{hidayati2017popularity} exploits an ensemble learning method to combine the outputs of an SVR and a CART (Classification And Regression Tree) models, previously trained to estimate the popularity score. The models have been trained by exploiting features extracted from user's information, image meta-data and visual aesthetic features extracted from the image. In particular, the authors take into account the post duration (i.e., the number of days the image was posted), the upload time, day and month.

\subsubsection{Popularity Dynamics}
Equation~\ref{eqPopScore} normalizes the number of interactions reached by an image by dividing the engagement measure by the time.
However, the measures $c_i$ related to social posts are cumulative values as they continuously collect the interactions between users and the social posts during their time on-line. Therefore, this normalization will penalize social media contents published in the past with respect to more recent contents, especially when the difference between the dates of posting is high. In other words, such a score does not take into account the evolution of the popularity over time. As consequence, two images with the same popularity dynamic could be ranked differently, depending on the download time.  
In particular, the most of the engagement obtained by a social media item is achieved in the first period, then the engagement measures become more stable. For example, the study presented in~\cite{valafar2009beyond} shows that photos obtain most of their engagement within the first 7 days since the date of upload. However, this study is focused on Flickr, and each social platform has its own mechanisms to show contents to users. There are very few works which takes into account the evolution of the image popularity over time (i.e., the dynamics of the image popularity).
The authors of~\cite{ortisACCESS8902084} presented a work on image popularity dynamic prediction. 
An important contribution is the definition of a dataset of $\sim$~20K Flickr images, crawled within the first 2 hours from posting, and their daily engagement scores for 30 days. The approach in~\cite{ortisACCESS8902084} considers two difference properties of a sequence: the sequence shape describes the trend of the popularity, whereas the sequence scale gives a degree of popularity in the whole period (i.e., the maximum value of engagement achieved after 30 days).
The authors performed a detailed analysis of the task, demonstrating that for a given sequence of popularity, the scale and shape properties are unrelated. Such a study defined a set of 50 typical shapes for the 30-days sequences that can be used as prototypes for the popularity dynamics, as well as a method to predict the shape and the scale of an image, given a set of post features. At test time, the two outputs are combined to infer the whole popularity sequence. Since the employed features are available at the moment of the post, this method is able to predict the whole sequence of popularity over 30 days with a daily granularity even before the image is posted. 

\subsection{Image Virality}
A recent emerging task, closely related to image popularity, is the prediction of the level of virality of an image.
The image virality is defined as the quality of a visual content~(i.e., images or videos) to be rapidly and widely spread on social networks~\cite{ricci2017virality}. 
Differently than popularity, the virality score takes into account also the number of resubmission of an image by different users. Therefore, images that became popular when they are posted, but not reposted, are not considered to be viral~\cite{deza2015understanding}.
These often involve images which content itself is less relevant, but are related to current events that drawn attention to the image in a specific period such as a flash news, or a tragedy.
The work in~\cite{ricci2017virality} focused on understanding the influence of image parts on its virality. In particular, the authors presented a method for the task of simultaneously detection and localicazion of virality in images. The detection consists on the prediction of the virality score on an image. The localization aims to detect which areas in an image are responsable for making the image viral, this allows to produce an heatmap which highlights the relevant areas of the input image.

\subsection{Relative Attributes}
As discussed in previous sections, several Visual Sentiment Analysis works aim to associate an image one sentiment label over a set of emotional categories or attributes. However, given a set of images that have been assigned to the same emotional category (e.g., joy), it would be interesting to determine their ranking with respect the specific attribute. 
Such a technique could suggest, for example, if a given image \textit{A} conveys more \virgolette{joy} than another image \textit{B}. For this purpose, several works on relative attributes can be exploited~\cite{parikh2011relative, altwaijry2013relative, fan2013relative, yu2015just}.
Furthermore, a ground truth dataset can be built by exploiting human annotators. Given a pair of images, the annotator is requested to indicate which image is closer to the attribute. In this way it's possible to obtain a proper ranking for each sentiment attribute.

\subsection{Common Sense}
With the aim to reduce the affective and cognitive gap between images and sentiments conveyed by them, we further need to encode the \virgolette{affective common-sense}. 
An Halloween picture 
can be classified as a negative image by an automatic system which considers the image semantics, however the knowledge of the context (i.e., Halloween) should affect the semantic concepts conveyed by the picture, hence its interpretation. 
This corresponds to the \virgolette{common-sense knowledge problem} in the field of knowledge representation, which is a sub-field of Artificial Intelligence.
Clearly, besides inferential capabilities, such an intelligent program needs a representation of the knowledge. 
By observing that is very difficult to build a Sentiment Analysis system that may be used in any context with accurate classification prediction, Agrawal et al.~\cite{agarwal2015sentiment} considered contextual information to determine the sentiment of text. Indeed, in this paper is proposed a model based on common-sense knowledge extracted from ConceptNet~\cite{liu2004conceptnet} ontology and context information. Although this work addresses the problem of Sentiment Analysis applied on textual data, as discussed above, the knowledge of the context related to what an image is depicting should affect its interpretation. Moreover, such results on textual analysis can be transferred to the visual counterpart. Furthermore, emerging approaches based on the Attention mechanism could be exploited to add such a context. The Attention mechanism is a recent trend in Deep Learning, it can be viewed as a method for making the Artificial Neural Network work better by letting the network know where to look as it is performing its task. For example, in the task of image captioning, the attention mechanism tells the network roughly which pixels to pay attention to when generating the text~\cite{xu2015show,you2016image}.

\subsection{Emoticon/Emoji}\label{secEmoji}

In this section we discuss about the possibility to exploit text ideograms, such emoticons and emoji, in the task of Sentiment Analysis on both visual and textual contents.
An emoticon is a textual shorthand that represents a facial expression. The emoticons have been introduced to allow the writer to express feelings and emotions with respect to a textual message. It helps to express the correct intent of a text sentence, improving the understanding of the message.
The emoticons are used to emulate visual cues in textual communications with the aim to express or explicitly clarify the writer's sentiment.
Indeed, in real conversations the sentiment can be inferred from visual cues such as facial expressions, pose and gestures. However, in textual based conversations, the visual cues are not present. 

The authors of~\cite{hogenboom2013exploiting} tried to understand if emoticons could be useful as well on the textual Sentiment Analysis task.
In particular, they investigated the role that emoticons play in conveying sentiment and how they can be exploited in the field of Sentiment Analysis. The authors manually labelled 574 emoticons as positive or negative, and combined this emoticon-lexicon with the text based Sentiment Analysis to perform document polarity classification considering both sentence and paragraph levels. 
A step further the emoticon, is represented by the emoji. An emoji is an ideogram representing concepts such as weather, celebration, food, animals, emotions, feelings, and activities, besides a large set of facial expressions.
They have been developed with the aim to allow more expressive messages. 
Emojis have become extremely popular in social media platforms and instant messaging systems. For example, in March 2015, Instagram  reported that almost half of the texts on its platform contain emojis~\cite{instagram2015}.

\begin{figure*}[t]
	\centering
	\includegraphics[width=0.7\linewidth]{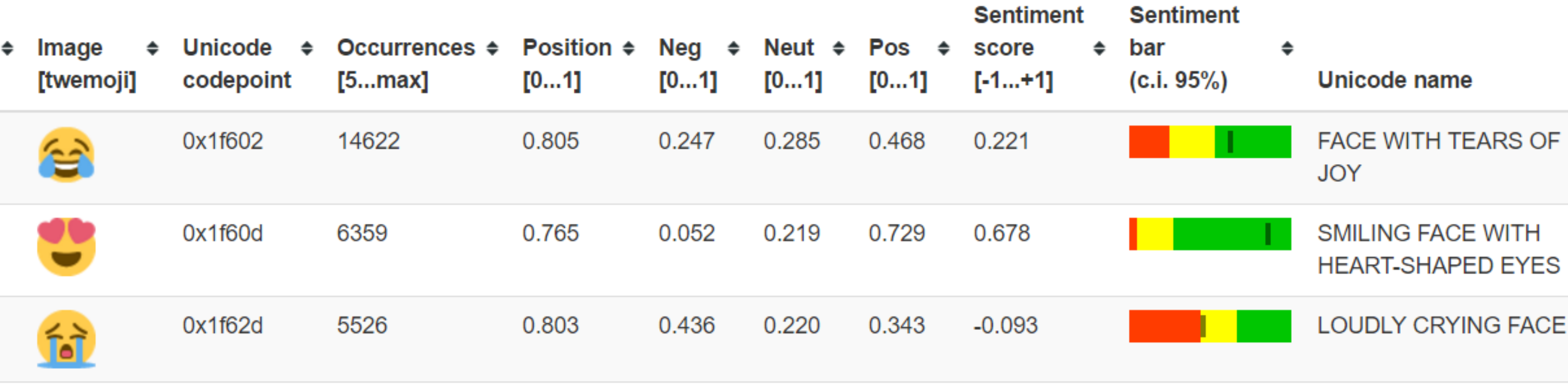}
	\caption{Some examples of the \textit{Emoji Sentiment Ranking} scores and statistics obtained by the study conducted in~\cite{novak2015sentiment}. The sentiment bar (10th column) shows the proportion of negativity, neutrality and positivity of the associated emoji.}
	\label{figEmojiranking}
\end{figure*}

In~\cite{cappallo2015image2emoji}, the authors exploited the expressiveness carried by emoji, to develop a system able to generate an image content description in terms of a set of emoji. The focus of this system is to use emoji as a means for image retrieval and exploration. Indeed, it allows to perform an image search by means of a emoji-based query.
This approach exploits the expressiveness conveyed by emoji, by leaning on the textual description of these ideograms (see the 10th column in Figure~\ref{figEmojiranking}). 
The work in~\cite{CappalloTEMP18} studied the ways in which emoji can be related to other common modalities such as text and images, in the context of multimedia research. This work also presents a new dataset that contains examples of both text-emoji and image-emoji relationships. Most of them contains also strong sentiment properties.

In~\cite{novak2015sentiment} the authors presented a sentiment emoji lexicon named Emoji Sentiment Ranking. In this paper, the sentiment properties of the emojis have been deeply analysed, and some interesting conclusions have been highlighted. For each emoji, the \textit{Emoji Sentiment Ranking} provides its associated positive, negative and neutral scores, represented by values between -1 and +1. The authors also proposed a visual tool, named sentiment bar, to better visualize the sentiment properties associated to each emoji (see Figure~\ref{figEmojiranking}).
The data considered in this analysis consists of 1.6 million labelled tweets.
This collection includes text written in 13 different languages. 
The authors found that the sentiment scores and ranking associated to emojis remain stable among different languages. This property is very useful to overcome the difficulties addressed in multilingual contexts.
This lexicon represents a precious resource for many useful applications~\footnote{The Emoji Sentiment Ranking scores computed by~\cite{novak2015sentiment} can be visualized at the following URL: http://kt.ijs.si/data/Emoji\_sentiment\_ranking/.}.

The work in~\cite{al2019smile} presented a dataset of 4M Twitter images with their associated emojis. Based on such large scale dataset, the authors defined a deep neural model for image-emoji embedding. The proposed system has been evaluated on the tasks of emoji prediction, sentiment analysis and emotion prediction. 

The results and the insights obtained in~\cite{cappallo2015image2emoji,CappalloTEMP18,al2019smile} and~\cite{novak2015sentiment} could be combined to exploit the sentiment conveyed by emoji on the task of Visual Sentiment Analysis, rather than exploiting textual information as in the most of the previous works. 
For instance, an image could be represented by considering the distribution of the associate emojis as a sentiment feature, taking a cue from the approach presented in~\cite{peng2015mixed}.

By a few years, Facebook has released a new \virgolette{reactions} feature, which allows users to interact with a Facebook post by using one of six emotional reactions (Like, Love, Haha, Wow, Sad, and Angry), instead of just having the option of \virgolette{liking} a post. These reactions corresponds to a meaningful subset of emoji. 



\section{Discussion} \label{discussion}
The papers presented in Section~\ref{stateoftheart} represent the most relevant works in the field fo Visual Sentiment Analysis aimed to infer the emotions evoked by pictures. As previously mentioned, there is not agreement on these works about meaningful input features, benchmark datasets nor emotion categories. For this reason, we further investigated these aspects in details in Section~\ref{problem}.
The basic polarity classification task considers two or three polarity levels (i.e., negative, neutral and positive). However, usually more sophisticated psychology based emotional models are used, such as the Plutchnik's wheel of emotions~\cite{plutchik1980general}, the Ekman's 5 basic emotions~\cite{ekman1987universals}, the 35 impression words~\cite{hayashi1998image}, or the Mikels' 8 basic emotions~\cite{mikels2005emotional}.
Moreover, the works that aim to infer a distribution over a set of sentiment categories~\cite{peng2015mixed,yang2017learning,ijcai2017-456} employ variations of Mikels' or Ekman's models.
Other works, employ specific categorizations such as the 5-scale sentiment categorization used in~\cite{xu2014visual}, or the ANP annotation introduced in~\cite{borth2013large}. 

The employed datasets can be organized into two main categories: psychological based or social media based.
Datasets based on psychological studies~\cite{lang1999international,yanulevskaya2008emotional,machajdik2010affective}, such as IAPS and GAPED, have different pros and cons. Indeed, the photos have been designed to select stimuli eliciting specific ranges of emotions, and the data can lean on reliable labelling~(i.e., human annotations supported by tools/scientists). However, this kind of dataset requires high efforts to be built and maintained over time. As consequence, is not possible to build large scale datasets. Moreover, considering the selection process of the photos, they results very different kind of images with respect to the ones shared through social media platforms.

More recent efforts have been spent in building large scale datasets of images shared through social media platforms~(e.g., Flickr, Twitter, Facebook, Instagram, etc.).
These media provide an easy way to retrieve images~(i.e., automatic crawling from social media platforms) at very large scale. Moreover, these images are often associated with accompanying metadata that provides additional context and help for the automatic labelling of the images~(e.g., by exploiting NLP on metadata).
However, such automatic labelling is often unreliable~(e.g., metadata includes sarcam, personal considerations, misleading text, etc.). 
Therefore, such large scale datasets need further filtering processes before their exploitation in Visual Sentiment Analysis systems.

Feature selection is crucial for any data analysis process. Data features are supposed to encode the information that the system will be able to infer.
There are three general approaches in the field of Image Sentiment Analysis: find correlations between visual features (e.g, colors, textures, etc..) and sentiment, learn image features from data (e.g., training a CNN), exploiting a multi-modal representation learning approach by combining visual and textual information through multimodal embedding systems (i.e., defining a new vector space).

\section{Summary and Future Directions}\label{conclusions}
In this paper we have discussed the main issues and techniques related to Visual Sentiment Analysis. The current state of the art has been analysed in detail, highlighting pros and cons of each approach and dataset. Although this task has been studied for years, the field is still in its infancy. Visual Sentiment Analysis is a challenging task due to a number of factors that have been discussed in this paper. 

The results discussed in this study, such as~\cite{machajdik2010affective}, agree that the semantic content has a great impact on the emotional influence of a picture. Images having similar colour histograms and textures could have completely different emotional impacts. 
As result, a representation of images which express both the appearance of the whole image and the intrinsic semantic of the viewed scene is needed. 
Early methods in the literature about Visual Sentiment Analysis tried to fill the so called \textit{affective gap} by designing visual representations. Some approaches build systems trained with human labelled datasets and try to predict the polarity of the images.
Other approaches compute the polarity of the text associated to the images~(e.g., post message, tags and comments) by exploiting common Sentiment Analysis systems that works on textual contents~\cite{esuli2006sentiwordnet, wilson2005recognizing}, and try to learn Machine Learning systems able to infer that polarity from the associated visual content.
These techniques have achieved interesting improvements in the tasks of image content recognition, automatic annotation and image retrieval~\cite{fu2014transductive, gong2014multi, gong2014improving,  guillaumin2010multimodal,   katsurai2014cross, li2015robust, rasiwasia2010new}.
However, it is impossible to know if such user provided text is related to the image content or to the sentiment it conveys. Moreover, the text associated to images is often noisy. Therefore, the exploitation of such text for the definition of either polarity ground truth or as an input source for a sentiment classifier have to address with not reliable text sources. A specific study on this issue is reported in~\cite{ortis2018visual}. 
Furthermore, some approaches exploit a combination of feature modalities~(often called views) to build feature space embeddings in which the correlation of the multi-modal features associated to the images that have the same polarity is maximized~\cite{katsurai2016image,ortis2018visual,campos2019sentiment,fortin2019multimodal}.
The results achieved by several discussed works suggest that exploiting multiple modalities is mandatory, since the sentiment evoked by a picture is affected by a combination of factors, beside the visual information.  This has been further confirmed by recent works that add the attention mechanism to the multi-modality approach~\cite{zhu2019joint, huang2019image}.
Studies in psychology and art theory suggested some visual features associated to emotions evoked by images. However, the most promising choice is given by representations automatically learned through neural networks, autoencoder and feature embedding methods. These approaches are able to find new feature spaces which capture contributes from the different input factor which the sentiment is affected by. The recent results in representation learning confirm this statement.


To this end, one important contribution is given by the availability of large and robust datasets. Indeed, in this study, we highlighted some issues related to the existing datasets.
Modern social media platforms allows the collection of huge amount of pictures with several correlated information. These can be exploited to define either input features and \virgolette{ground truth}. However, as highlighted before, these textual information need to be properly filtered and processed, in order to avoid the association of noisy information to the images.

Systems with broader ambitions could be developed to address the new challenges~(e.g., relative attributes, popularity prediction, common-sense, etc.) or to focus on new emerging tasks~(e.g., image popularity prediction, sentiment over time, sentiment by exploiting ideograms, etc.).
For instance, ideograms help people to reduce the gap between real and virtual communications. Thanks to the diffusion of social media platforms, the use of emojis has been growing for years, and they are now integrated to the way people communicate in the digital world. They are commonly used to express user reactions with respect to messages, pictures, or news. Thus, the analysis of such new communication media could help to improve the current state of the art performances.  

This paper aimed to give a comprehensive overview of the Visual Sentiment Analysis problem, the relative issues, and the algorithms proposed in the state of the art. 
Relevant points with practical applications in business fields which would benefit from studies in Sentiment Analysis on visual contents have been also discussed. 
This survey addresses all the aspects related to Visual Sentiment Analysis, and for each of these, relevant references, datasets and results are presented. It also proposes a critical perspective for each considered issue. 
The paper aims to serve as a starting point for researchers that want to tackle tasks in the field of Visual Sentiment Analysis and related challenges, as it is presented as a structured and critique review of previous works, description of the available datasets, features and techniques.  
It also aims to clear the way for new solutions, by suggesting new and interesting techniques and sources of information that can be explored to tackle the problem.
\\
Although the psychological basis~(e.g., emotional models) and some techniques are often applied on both images and videos, the presented study is mostly devoted to the analysis of still images. Future extensions of this paper could be devoted to the review of the state of the art in Visual Sentiment Analysis applied on videos. In the case of videos, the temporal dimension and the presence of the audio or other type of temporal signals~(e.g., EEG~\cite{liu2017real}) pave the way for a number of sentiment related tasks on visual contents such as multi-modal continuous emotional state recognition~\cite{liu2017real} or scene popularity~\cite{ortis2015recfusion,battiato2017organizing}, among others. Several datasets designed for these tasks are already available~\cite{liu2017real,ortis2015recfusion,battiato2017organizing}.

\bibliographystyle{unsrt}  
\bibliography{ms}  

%
%
%
%

\end{document}